\documentclass[11pt]{article}

\usepackage[margin=1in]{geometry}
\usepackage[T1]{fontenc}
\usepackage[utf8]{inputenc}
\usepackage{textcomp}
\usepackage{amsmath,amssymb,amsfonts}
\usepackage{graphicx}
\usepackage{booktabs}
\usepackage{multirow}
\usepackage{array}
\usepackage{tabularx}
\usepackage{threeparttable}
\usepackage{xcolor}
\usepackage[numbers,sort&compress]{natbib}
\usepackage[hidelinks]{hyperref}

\newcommand{\rowstrut}{\rule{0pt}{2.4ex}}
\newcommand{\imp}[1]{\textcolor{blue}{#1}}
\newcommand{\impPos}[1]{\textcolor{blue}{#1}}
\newcommand{\impNeg}[1]{\textcolor{red}{#1}}

\begin{document}

\title{Evaluating Time-Series Foundation Models and Multimodal Dietary Context for CGM Forecasting}
\author{Bowen Zhang$^{1,\dagger}$, Hsiu-Wen Cheng$^{1,\dagger}$,
Hongyu Yang$^{1,\dagger}$, Evie L. Shen$^{2}$,\\[0.25em]
Joleen Vansomphone$^{3}$, Yuna Li$^{4}$, Kerry Zhou$^{5}$,
Zitian Qu$^{6}$,\\[0.25em]
Suning Zhao$^{1}$, Xiangning Deng$^{1}$, Hua Zhou$^{1,*}$,
Jin J. Zhou$^{1,*}$}

\date{}

\maketitle

\begin{center}
\begin{minipage}{0.96\textwidth}
\small
$^{1}$University of California, Los Angeles, Los Angeles, CA, U.S.A.\\
$^{2}$Union County Magnet High School, Scotch Plain, NJ, U.S.A.\\
$^{3}$Huntington Beach High School, Huntington Beach, CA, U.S.A.\\
$^{4}$Crean Lutheran High School, Irvine, CA, U.S.A.\\
$^{5}$Portola High School, Irvine, CA, U.S.A.\\
$^{6}$Tsinghua University, Beijing, China.\\[0.35em]
$^{\dagger}$Bowen Zhang, Hsiu-Wen Cheng and Hongyu Yang contributed equally to this research.\\
$^{*}$Corresponding authors: Hua Zhou and Jin J. Zhou.
\end{minipage}
\end{center}

\begin{center}
\begin{minipage}{0.96\textwidth}
\textit{This work has been submitted to the IEEE for possible publication. Copyright may be
transferred without notice, after which this version may no longer be accessible.}
\end{minipage}
\end{center}

\normalfont\normalsize
\begin{abstract}
    Continuous glucose monitoring (CGM) provides high-frequency measurements of glucose dynamics and enables short-term glucose forecasting for diabetes management. Although time-series foundation models have shown strong general forecasting ability, their effectiveness for CGM prediction and the added value of multimodal dietary context remain unclear. We conduct a comprehensive empirical study using eight public CGM datasets spanning Type 1 diabetes, Type 2 diabetes, and non-diabetes populations. Under a unified protocol across multiple context lengths and prediction horizons, zero-shot foundation models did not consistently outperform strong task-specific baselines such as Elastic Net and PatchTST. In contrast, lightweight fine-tuning substantially improved forecasting performance. For example, fine-tuned Chronos-Bolt reduced RMSE by 6.5\%–18.4\% in the T1D cohort and by 8.6\%–18.2\% in the non-diabetes/T2D cohort, with comparable improvements in both in-distribution and out-of-distribution test settings. We further evaluate multimodal dietary context using CGMacros, which provides temporally aligned CGM signals, food images, and macronutrient records. A residual-based fusion framework reduced overall RMSE by approximately 3\% and postprandial RMSE by approximately 15\% relative to the CGM-only baseline. Moreover, Chronos-based CGM representations were more strongly correlated with observed postprandial glucose increments than representations from LSTM and CatBoost, even after those models incorporated additional dietary modalities, suggesting that pretrained temporal representations better preserve meal-induced excursion patterns. These findings show that foundation models require CGM-specific adaptation for reliable forecasting and that dietary context provides clinically meaningful signals beyond CGM alone, especially during postprandial periods.

\end{abstract}

\noindent\textbf{Keywords:} Continuous glucose monitoring, Food image representation, Foundation models, Multimodal learning, Time-series forecasting.

\section{Introduction}
\label{sec:introduction}

Diabetes mellitus (DM) is a chronic metabolic disease characterized by persistently elevated blood glucose levels due to impaired insulin secretion or insulin resistance, leading to progressive organ damage over time \cite{Hossain2024DiabetesPublicHealth}. Both Type~1 diabetes (T1D) and Type~2 diabetes (T2D) populations face substantial challenges in maintaining stable glycemic control. Continuous glucose monitoring (CGM), which provides high-frequency measurements of interstitial glucose, enables detailed characterization of glycemic dynamics and has been shown to capture glycemic profiles associated with increased mortality risk and complications \cite{OkunoContinuous2025, 10.1210/clinem/dgac692, Tomoki_association_2026, ReavenInitation_2025}. Large-scale real-world studies integrating CGM device data with electronic health records further demonstrate substantial heterogeneity in CGM usage patterns and glycemic control metrics across T1D and T2D populations, underscoring the need for data-driven modeling of glucose dynamics \cite{Okuno2024Assessing,Okuno2025Temporal}. Beyond enabling descriptive characterization of glucose dynamics, CGM initiation has also been linked to improved glycemic control and fewer adverse clinical events, providing strong clinical motivation for developing models that can leverage CGM data to inform glucose management \cite{ReavenInitiation_2023}.

Early CGM-based glucose forecasting relied on classical statistical and machine learning approaches, including linear regression, ARIMA\cite{Yang2019AdaptiveARIMA,Hyndman2008AutoARIMA}, and Elastic Net \cite{Xie2020BenchmarkingBG}. Although computationally efficient and interpretable, these classical approaches are often limited in their ability to capture complex temporal dependencies and to scale to longer prediction horizons. More recently, deep learning methods, particularly Transformer-based architectures, have demonstrated improved performance by modeling long-range temporal dependencies and multi-scale glucose dynamics \cite{Lim2021TFT,sergazinov2023gluformer}, outperforming earlier convolutional and recurrent models \cite{Mhaskar2017DeepBG,Song2019EMDLSTM,Sun2018LSTM}.

Beyond task-specific architectures, the field has increasingly turned to large-scale time-series foundation models pretrained on diverse and heterogeneous temporal datasets \cite{rasul2024lagllamafoundationmodelsprobabilistic, ansari2024chronoslearninglanguagetime, ansari2025chronos2univariateuniversalforecasting, das2024decoderonlyfoundationmodeltimeseries}. These models exhibit strong cross-domain generalization, suggesting that pretrained temporal representations may transfer effectively to clinically relevant forecasting tasks. However, it remains unclear whether such benefits persist for CGM forecasting under clinically realistic settings involving limited historical context, varying prediction horizons, and heterogeneous patient populations, or whether zero-shot deployment is sufficient relative to lightweight fine-tuning.

At the same time, glucose dynamics are strongly influenced by external physiological and behavioral factors that are not fully observable from CGM signals alone. Dietary intake, in particular, induces abrupt and heterogeneous perturbations that challenge univariate forecasting approaches and has motivated multimodal models that integrate CGM with complementary contextual information such as wearable sensors or dietary records \cite{Chowdhury2024MMGNet}. Despite encouraging results, it remains unclear which aspects of dietary context are most informative for postprandial glucose dynamics, and whether visually inferred dietary representations provide complementary value beyond structured nutritional variables.

GlucoBench represents an important step toward standardized evaluation in CGM forecasting by curating public datasets and defining unified benchmarking protocols \cite{Sergazinov2024GlucoBench}. Building on this foundation, we conduct a two-part empirical study to clarify the roles of pretrained temporal representations and dietary multimodality in CGM forecasting. First, using curated public CGM datasets, we evaluate classical statistical models, deep learning baselines, and recent time-series foundation models across historical context lengths, prediction horizons, and populations, explicitly comparing zero-shot and lightweight fine-tuning. Second, leveraging the CGMacros dataset \cite{Das2025CGMacros}, which provides temporally aligned CGM signals, food images, and macronutrient records, we examine the contribution of multimodal dietary context using a carefully designed residual-based fusion framework. By separating the CGM-only baseline prediction from a meal-driven residual correction branch, this design enables controlled comparisons across dietary input modalities and helps assess the complementary roles of visual and nutritional signals.

Together, this study provides a systematic evaluation of time-series foundation models and multimodal dietary information for CGM forecasting under clinically realistic settings. Through unified benchmarking across public CGM datasets and controlled multimodal experiments using CGMacros, this work examines the extent to which pretrained temporal representations and dietary context can support practical glucose forecasting. Code for data preprocessing, model training, and evaluation are available at \url{https://github.com/BowenZhang2001/CGM_Forecasting.git}.

\section{Related work}

Prior work on glucose forecasting spans a wide range of modeling paradigms, data sources, and experimental setups, reflecting diverse clinical contexts and research objectives in this domain. Existing studies have explored approaches based on classical statistical models and deep learning architectures on CGM or related time-series data \cite{Lutsker2026FoundationCGM, Luo2024CGMFoundation, Gluformer}, as well as privacy-preserving and federated learning frameworks for cross-patient blood glucose prediction \cite{Federated,meta}. In parallel, multimodal inputs have been incorporated to provide contextual information for glucose prediction, such as dietary intake and physiological signals \cite{Plis2014BGPrediction, Wu2026PPGRModifiableFactors}. However, many of these efforts are evaluated under task-specific designs tailored to particular cohorts or objectives, which complicates direct comparison.

Within multimodal glucose modeling, a substantial body of work incorporates dietary factors, including macronutrients and physical activity \cite{Hotta2024DietExercise}, insulin dosing and blood biomarkers \cite{Xiong2024PersonalizedPPG}, or manually logged meal features \cite{Brugger2025PPG}. Many of these studies characterize postprandial glycemic response through summary outcomes or short-term response metrics \cite{Pustozerov2020GDM}, rather than framing postprandial dynamics as a multi-horizon temporal forecasting problem. As a result, multimodal models are often developed and evaluated using heterogeneous formulations and datasets, which hinders direct comparison of their individual contributions. \cite{Wolber2025MultimodalLLM, Sergazinov2024GlucoBench}.

More broadly, many multimodal approaches in the glucose modeling literature emphasize dietary assessment objectives, such as calorie or intake estimation, and adopt feature-level \cite{Plis2014BGPrediction, Zecchin2016InsulinMeal}, or mechanistic pipelines instead of end-to-end temporal forecasting models. Despite recent advances in time-series foundation models, their use in CGM forecasting has received limited systematic evaluation, particularly with respect to comparisons between zero-shot and lightweight fine-tuning under consistent forecasting horizons. Taken together, these lines of work suggest that the relative roles of pretrained temporal representations and dietary modalities in short-horizon glucose forecasting remain insufficiently understood under controlled and comparable settings.

\section{Data}

\subsection{Ethics statement}

This study used publicly available, de-identified datasets. All data were obtained from previously published studies or public repositories, and no new data were collected from human participants. Informed consent and ethical approval were obtained in the original studies, as applicable. Therefore, no additional informed consent was required for the present secondary analysis.

\subsection{Description}

We considered eight publicly available CGM datasets collected across diverse study designs and populations. These datasets include three cohorts of individuals with T1D \cite{Anderson2016ClosedLoop,Brown2019ClosedLoop,Lynch2022InsulinOnlyBP}, one cohort of individuals with T2D \cite{Broll2021iglu}, one non-diabetes cohort \cite{Shah2019CGMHealthy}, and three mixed cohorts comprising non-diabetes, pre-diabetes, and T2D participants \cite{Das2025CGMacros, Hall2018Glucotypes, Colas2019DFA}.

Given the limited availability of large-scale public T2D CGM datasets and the data demands associated with training and evaluating time-series foundation models, we grouped the three T1D datasets into a single T1D cohort and combined the remaining five datasets into a non-diabetes/T2D cohort. This grouping strategy was applied consistently throughout the forecasting benchmarks to support stable estimation and facilitate comparable evaluation across models.

For multimodal forecasting experiments, we focused exclusively on the CGMacros dataset, which uniquely provides temporally aligned multimodal records, including CGM measurements, food images, and structured macronutrient records \cite{Das2025CGMacros}. This alignment enables us to isolate and evaluate the incremental contribution of dietary context to glucose forecasting performance.

\begin{figure}[htbp]
  \centering
  \includegraphics[width=\linewidth]{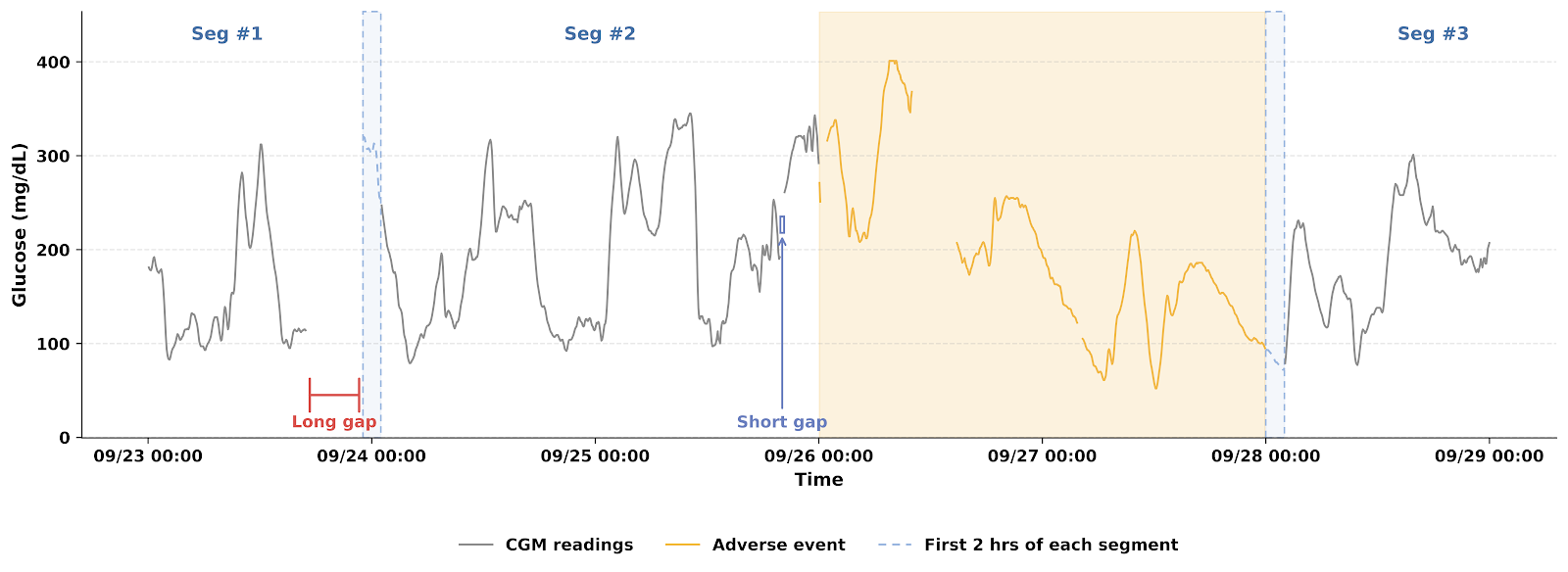}
  \caption{CGM interpolation and segmentation. CGM readings were partitioned into segments separated by long gaps exceeding 2 hours, which commonly arise from sensor replacement or prolonged data loss. Periods corresponding to adverse events were excluded. To mitigate potential instability at segment boundaries, the first two hours of each segment were removed.}
  \label{fig:preprocessing}
\end{figure}

\subsection{Preprocessing}
\label{sec:preprocess}

CGM data were preprocessed using a standardized pipeline to ensure temporal consistency, prevent information leakage, and support comparable evaluation across datasets. The preprocessing pipeline consisted of the following steps:

\textbf{Cohort selection.} Analyses were restricted to adult participants. For T1D cohorts, we retained only CGM observations collected before randomization and outside closed-loop control periods, because the source studies were designed around closed-loop insulin delivery systems and post-randomization data may reflect algorithm-mediated glucose regulation rather than user-managed glycemic control. Implausible glucose values and initial unstable sensor readings were also removed.

\textbf{Interpolation and segmentation.} CGM trajectories were mapped to a uniform temporal grid. Short missing intervals of less than 2~hours were linearly interpolated, whereas longer gaps were used to split trajectories into independent temporal segments to avoid artificial extrapolation across extended missing periods. An overview of the gap-handling and segmentation strategy is illustrated in Figure~\ref{fig:preprocessing}.

\textbf{Data splitting.} For model evaluation, a subject-level out-of-distribution (OOD) split was adopted by holding out 20\% of participants as an external test cohort. The remaining data were divided into training, validation, and in-distribution (ID) test sets using chronological splits to prevent information leakage across time. For multimodal experiments on CGMacros, subject-level OOD splits were not applied due to the limited number of participants. Instead, strictly chronological training, validation, and test splits were used.

\textbf{Multimodal alignment.} For multimodal experiments, dietary images and nutritional records were temporally aligned with CGM measurements on a uniform 5-minute grid.

As shown in Table~\ref{tab:dataset}, after preprocessing, the forecasting comparison cohort comprised 145 participants with T1D, contributing approximately 0.64 million CGM observations, and 404 participants without diabetes or with T2D, contributing approximately 0.59 million observations. For the multimodal forecasting experiments, the CGMacros dataset comprised 44 participants, with approximately 125,000 CGM observations and 1,611 meal records after preprocessing.

\begin{table}[htbp]
\caption{Overview of public CGM datasets used for model training and evaluation.}
\label{tab:dataset}
\centering
\renewcommand{\arraystretch}{0.7}
\scalebox{0.85}{
\begin{threeparttable}
\begin{tabular}{lccccc}
\toprule
\multirow{2}{*}{Dataset}
& \multirow{2}{*}{DM Type}
& \multirow{2}{*}{Device}
& \multicolumn{2}{c}{\# of subjects}
& \multirow{2}{*}{\begin{tabular}[c]{@{}c@{}}Mean length of records\\ (days, post-processing)\end{tabular}} \\
\cmidrule(lr){4-5}
& & & Raw & Processed & \\
\midrule
Anderson et al. (2016)~\cite{Anderson2016ClosedLoop} & T1D & Dexcom G4 & 30  & 29   & 25.5 \\
Broll et al. (2021)~\cite{Broll2021iglu}   & T2D & Dexcom G4 & 5   & 5    & 9.5  \\
Brown et al. (2019)~\cite{Brown2019ClosedLoop}   & T1D & Dexcom G6 & 168 & 95   & 12.2 \\
CGMacros (2025)~\cite{Das2025CGMacros}       & Non-diabetes, T2D & Dexcom G6 Pro & 45 & 45\textsuperscript{*} & 9.7 \\
Colas et al. (2019)~\cite{Colas2019DFA}   & Non-diabetes, T2D & MiniMed iPro & 191 & 191 & 1.9 \\
Hall et al. (2018)~\cite{Hall2018Glucotypes}    & Non-diabetes, T2D & Dexcom G4 & 57 & 57 & 5.8 \\
Lynch et al. (2022)~\cite{Lynch2022InsulinOnlyBP}   & T1D & Dexcom G6 & 90 & 21 & 15.3 \\
Shah et al. (2019)~\cite{Shah2019CGMHealthy}    & Non-diabetes & Dexcom G6 & 153 & 106 & 8.1 \\
\bottomrule
\end{tabular}

\begin{tablenotes}[flushleft]
\item \textsuperscript{*} In the multimodal forecasting, one subject with insufficient or irregular data was excluded.
\end{tablenotes}

\end{threeparttable}
}
\end{table}

\section{Experiment}

\subsection{Evaluation of Time-Series Foundation Models for Glucose Forecasting}

\subsubsection{Models and forecasting tasks}

We benchmarked forecasting performance across multiple historical context lengths and prediction horizons using a diverse set of classical baselines, deep learning models, and pretrained time-series foundation models.

As simple baselines, we included \textbf{Last Observation Carried Forward (LOCF)}, \textbf{AutoARIMA}~\cite{Hyndman2008AutoARIMA}, and \textbf{Elastic Net}~\cite{Zou2005ElasticNet}. Among deep learning baselines, we evaluated a \textbf{Long Short-Term Memory (LSTM)} network~\cite{LSTM} and \textbf{PatchTST}~\cite{Nie2023PatchTST}, a Transformer-based architecture designed for long-horizon time-series forecasting. We further considered four pretrained time-series foundation models: \textbf{Chronos-Bolt-Tiny} and \textbf{Chronos-Bolt-Mini}, lightweight encoder–decoder Transformers that differ primarily in model capacity~\cite{ansari2024chronoslearninglanguagetime}; \textbf{Chronos2-Small}, a compact pretrained Transformer from the Chronos family that operates directly in the continuous value space~\cite{ansari2025chronos2univariateuniversalforecasting}; and \textbf{TimesFM~2.5}, a decoder-only model optimized for multi-horizon probabilistic forecasting~\cite{das2024decoderonlyfoundationmodeltimeseries}.

Foundation models were evaluated under two usage regimes: \textit{zero-shot} inference, where pretrained weights were applied without task-specific adaptation, and \textit{fine-tuning} using the CGM training data. All models were evaluated across multiple input–output configurations, varying the historical context length (4~h, 12~h, and 24~h) and prediction horizon (30~min, 1~h, and 2~h). Each context–horizon combination was treated as an independent forecasting task, with non-pretrained models trained and foundation models fine-tuned separately when applicable.

\subsubsection{Experimental protocol}

Datasets were split into training, validation, ID test, and OOD test sets following the protocol in Section~\ref{sec:preprocess}. Forecasting instances were constructed using a sliding window, where historical contexts of varying lengths were used to predict future glucose trajectories. A fixed stride of 24 time steps was used for both training and evaluation across all models.

Hyperparameters for deep learning models were selected via hyperparameter optimization (HPO) with 10 trials per configuration. For each model and context--horizon setting, the configuration achieving the best validation performance was used for evaluation on the ID and OOD test sets.

To account for training stochasticity, LSTM, PatchTST, and all fine-tuned foundation models were evaluated over multiple random seeds. LSTM, PatchTST, and Chronos-Bolt models were repeated with 10 seeds, while Chronos2-Small and TimesFM 2.5 were repeated with 5 seeds due to substantially higher computational cost. Mean performance across repetitions is reported. In contrast, zero-shot foundation models and classical baselines were evaluated using a single run, as no stochastic training is involved.

Given a CGM time series $\{ x^{(i)}_t \}_{t=1}^{T_i}$ from subject $i$, at each forecasting time $t$ the model predicts future values
$\hat{\mathbf{x}}^{(i)}_{t+1:t+H}=(\hat{x}^{(i)}_{t+1}, \ldots, \hat{x}^{(i)}_{t+H})$ based on a historical context of length $L$. Performance is evaluated using root mean squared error (RMSE):
$$
\mathrm{RMSE}^{(i)}_t = \sqrt{\frac{1}{H} \sum_{h=1}^{H} \left(x^{(i)}_{t+h} - \hat{x}^{(i)}_{t+h}\right)^2},
$$
and final metrics are obtained by averaging over all forecasting instances across subjects in the test set:

$$
\mathrm{RMSE} = \frac{1}{|\mathcal{T}_{\mathrm{test}}|} \sum_{(i,t) \in \mathcal{T}_{\mathrm{test}}} \mathrm{RMSE}^{(i)}_t.
$$

\subsection{Multimodal Integration for Glucose Forecasting}

\subsubsection{Visual backbone for food image representation}

To incorporate food image information into multimodal glucose forecasting, we evaluated several visual backbone architectures on the Food-101 dataset~\cite{bossard2014food101}. All models were fine-tuned end-to-end, and the backbone with the best validation accuracy was selected to extract food image embeddings for downstream multimodal CGM forecasting on the CGMacros dataset.

Food-101 classification accuracy was used as a proxy for representation quality. Using the selected backbone architecture, we further compared two supervision strategies during Food-101 training. In the first setting, models were trained using standard 101-class food category labels. In the second setting, models were trained using nutrition-aligned supervision, formulated as a multi-output regression task in which the targets were category-level nutritional profiles, including calories and macronutrients, obtained from a publicly available Food-101 nutrition dataset~\cite{food101_nutritional_info_kaggle}.

These two training paradigms were compared based on downstream CGM forecasting performance, allowing us to assess whether nutrition-aligned visual representations provide additional benefit beyond category-supervised embeddings.

\begin{figure}[htbp]
  \centering
  \includegraphics[width=0.8\linewidth]{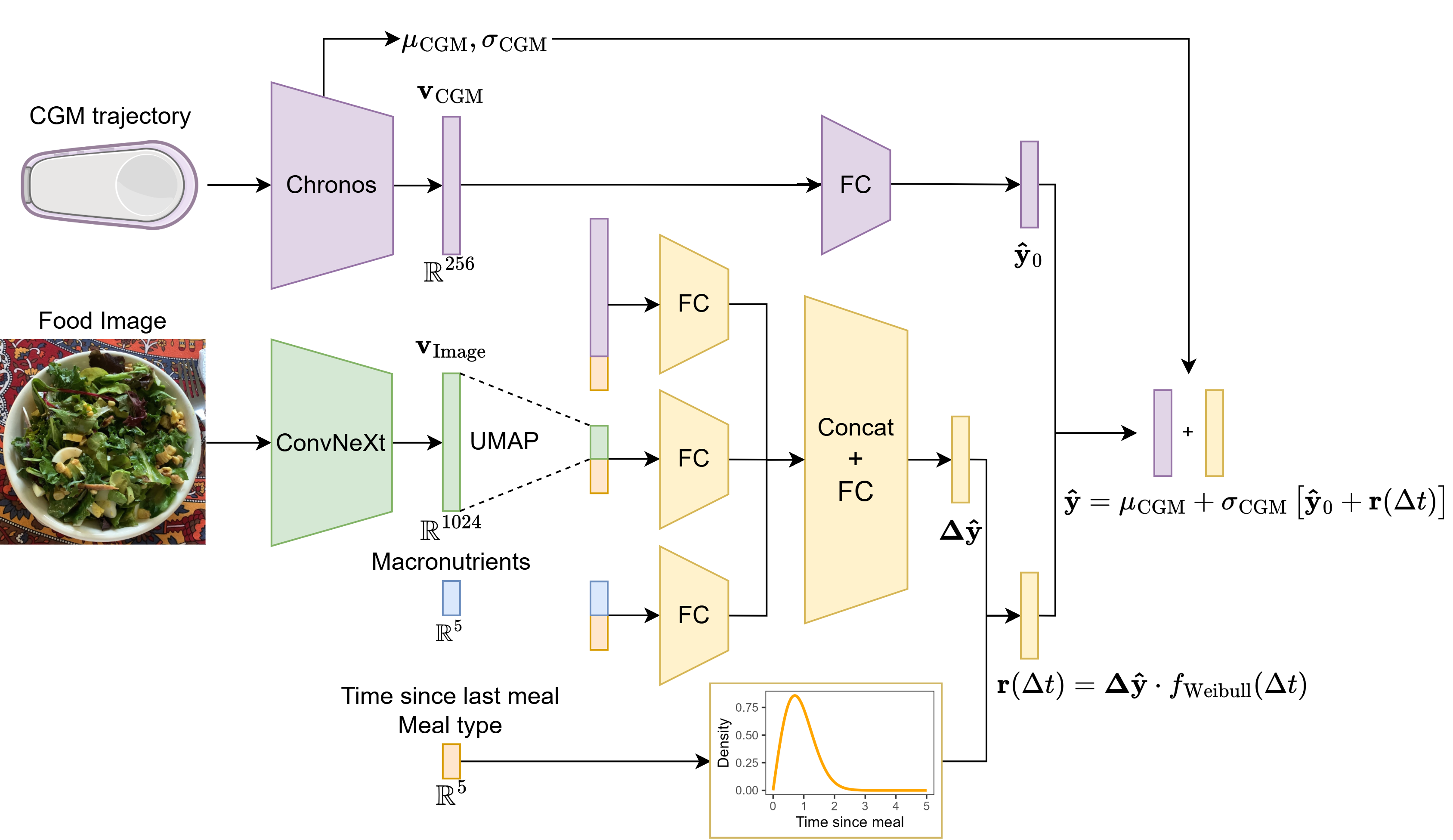}
  \caption{Overview of the multimodal CGM forecasting framework. CGM time-series are encoded by a Chronos backbone to produce a baseline prediction $\hat {\mathbf{y}}_0$. Multimodal meal information is used to estimate a residual effect $\Delta \hat {\mathbf{y}}$ which is temporally modulated by a Weibull-based gating function and added to the baseline to obtain the final glucose forecast.}
  \label{fig:fusion_arch}
\end{figure}

\subsubsection{Multimodal CGM forecasting with food image and macronutrients}

Figure~\ref{fig:fusion_arch} illustrates the proposed multimodal CGM forecasting framework, which combines a CGM-only Chronos backbone with a meal-driven residual correction module. The CGM trajectory is first passed through a fine-tuned Chronos-Bolt-Tiny model to obtain a normalized baseline forecast $\hat{y}_0$ and an intermediate CGM representation $\mathbf{v}_{\mathrm{CGM}}$. The CGM-specific mean and standard deviation, $\mu_{\mathrm{CGM}}$ and $\sigma_{\mathrm{CGM}}$, are retained for transforming predictions back to the original scale. In all multimodal experiments, Chronos is used as a fixed CGM encoder so that the contribution of meal-related information can be isolated through the residual branch.

Meal information is represented using food images, structured nutritional variables, and meal timing/type features. Food images are encoded using the selected ConvNeXt backbone to obtain high-dimensional embeddings $\mathbf{v}_{\mathrm{Image}}$, which are then projected to an 8-dimensional latent space using UMAP to reduce model complexity while preserving local neighborhood structure. The reduced image embeddings and macronutrient features, including carbohydrates, protein, fat, fiber, and total caloric content, are separately transformed through fully connected layers and then combined to estimate a residual correction magnitude $\Delta \hat{y}$.

To account for the time-varying postprandial effect of meals, the residual correction is modulated by a Weibull-based temporal gate parameterized by the elapsed time since the most recent meal, $\Delta t$. The final residual is defined as
$$
r(\Delta t) = \Delta \hat{y} \cdot f_{\mathrm{Weibull}}(\Delta t),
$$
where the gate allows meal-related information to contribute most during the postprandial period and decay at longer delays. The final glucose prediction is obtained by adding this residual correction to the CGM-only baseline forecast and transforming back to the original glucose scale:
$$
\hat{y}=\mu_{\mathrm{CGM}}
+
\sigma_{\mathrm{CGM}}
\left[
\hat{y}_0 + r(\Delta t)
\right].
$$
This residual design separates baseline temporal prediction from meal-related correction, enabling controlled comparisons of image, nutritional, and timing information in multimodal CGM forecasting.

\subsubsection{Ablation study}

We conducted ablation studies to quantify the contribution of individual modalities and the supervision strategy used to learn food image embeddings in the proposed multimodal CGM forecasting framework. In all ablations, the CGM backbone was kept fixed, with an input context length of 24 hours and a forecasting horizon of 2 hours.

Hyperparameters were selected based on validation performance and then fixed. Each configuration was trained and evaluated over 10 independent runs, with performance reported as the mean RMSE.

\paragraph{Effect of modality integration} We first evaluated the incremental benefit of incorporating meal-related information by comparing four model configurations:
\begin{enumerate}
  \item CGM only without the residual correction branch;
  \item CGM with food image embeddings;
  \item CGM with macronutrient features;
  \item Full multimodal model combining CGM, food images, and macronutrients.
\end{enumerate}

This comparison isolates the individual and joint contributions of visual and nutritional modalities, and assesses whether their combination yields complementary gains.

\paragraph{Effect of image supervision strategy} We next evaluated the impact of visual supervision by comparing food image embeddings trained using standard food category labels with those trained using nutrition-aligned supervision. All other model components and training procedures were held constant. This analysis tests whether aligning visual representations with nutritionally relevant properties helps downstream glucose forecasting.

Together, these ablations assess the contribution of each modality and design choice in the proposed multimodal framework.

\subsubsection{Comparison with alternative CGM encoders}

To examine whether the contribution of meal-related modalities generalizes beyond a Chronos-based CGM representation, we performed two complementary comparisons. First, in a CGM-encoder swap, Chronos CGM embeddings were replaced with LSTM embeddings while keeping the multimodal fusion and residual correction modules unchanged. This comparison evaluates alternative sequence representations under the same multimodal architecture. Second, as a reference baseline, we included a standalone CatBoost model trained on windowed raw CGM inputs, without using the proposed fusion framework.

The LSTM-based variant was used as a component-level comparison because it preserves the same fusion and residual correction structure while changing only the CGM encoder. In contrast, CatBoost relies on a tree-based formulation applied to raw CGM input windows and does not impose an explicit sequential architecture. Given these different modeling assumptions, CatBoost was included as a reference baseline.

All models were evaluated under the same data splits and evaluation protocol. In addition to overall and postprandial forecasting accuracy measured by RMSE, we evaluated each model's ability to capture postprandial glucose excursions using the postprandial glucose increment (PGI). Predicted PGI was defined as the maximum predicted glucose value within the 2-hour forecasting horizon minus the pre-meal glucose level, and observed PGI was defined analogously using the observed glucose trajectory. We reported the correlation between predicted and observed PGI to assess whether each model captured meal-induced glucose increases.

These comparisons assess whether the contribution of meal-related modalities is consistent across CGM representations and whether this contribution extends to the characterization of postprandial glucose excursions.

\section{Results}

\subsection{CGM-Only Forecasting Performance of Foundation Time-Series Models}

\subsubsection{Zero-shot foundation models do not consistently outperform strong baselines}

Table~\ref{tab:cgm_perf_RMSE_ID_combined} reports comprehensive CGM forecasting results across three historical context lengths (4, 12, and 24 hours) and three prediction horizons (30 minutes, 1 hour, and 2 hours) for both the T1D and Non-diabetes/T2D cohorts. All models are evaluated on the ID test set.

Across both cohorts, zero-shot foundation models do not consistently outperform strong classical or task-specific baselines, even with extended historical context. Elastic Net and PatchTST remain highly competitive, particularly for short-term forecasting where strong temporal autocorrelation benefits simpler models.  

Among zero-shot foundation models, performance varies substantially by architecture. The Chronos-Bolt variants (Tiny and Mini) frequently exhibit higher RMSE; for example, in the T1D cohort with a 24-hour context, they underperform Elastic Net and PatchTST by approximately 7--10\% at the 30-minute horizon, with persistent gaps at longer horizons.  

Larger foundation models such as TimesFM-2.5 and Chronos2-small achieve more competitive zero-shot results but fail to deliver systematic improvements across cohorts and settings. Overall, these findings indicate that off-the-shelf pretrained time-series models alone may be insufficient for reliable CGM forecasting without task-specific adaptation.

\begin{table}[htbp]
\centering

% Caption and label are outside the scaling scope.
\caption{CGM forecasting RMSE performance (mean $\pm$ SD) on the ID test set for T1D and Non-diabetes/T2D cohorts.}
\label{tab:cgm_perf_RMSE_ID_combined}

% Only the table content is scaled.
\begingroup
\renewcommand{\arraystretch}{0.4}
\setlength{\tabcolsep}{3.3pt}
\small
\scalebox{0.62}{%
\begin{tabular}{
  l
  l
  ccc ccc ccc
}
\toprule
\multirow{2}{*}{} &
\multirow{2}{*}{\textbf{Model}} &
\multicolumn{3}{c}{Ctx: 4-hr} &
\multicolumn{3}{c}{Ctx: 12-hr} &
\multicolumn{3}{c}{Ctx: 24-hr} \\
\cmidrule(lr){3-5}
\cmidrule(lr){6-8}
\cmidrule(lr){9-11}
& & 30-min & 1-hr & 2-hr
  & 30-min & 1-hr & 2-hr
  & 30-min & 1-hr & 2-hr \\
\midrule

% ========================= T1D (RMSE) =========================
\multirow{13}{*}[-10pt]{%
  \rotatebox{90}{\textbf{T1D\qquad\qquad}}%
}
& LOCF
& 12.70 & 20.96 & 33.69
& 12.70 & 20.96 & 33.69
& 12.70 & 20.96 & 33.69 \\

& AutoARIMA
& 11.86 & 22.85 & 46.00
& 10.95 & 20.26 & 37.70
& 10.59 & 18.88 & 33.87 \\

& Elastic Net
& 10.36 & 18.32 & 30.60
& 10.28 & 18.21 & 30.66
& 10.23 & 17.98 & 30.09 \\

& LSTM
& 10.28 $\pm$ 0.09 & 18.11 $\pm$ 0.11 & 30.28 $\pm$ 0.09
& 10.25 $\pm$ 0.09 & 18.07 $\pm$ 0.10 & 30.42 $\pm$ 0.17
& 10.37 $\pm$ 0.10 & 18.30 $\pm$ 0.24 & 30.05 $\pm$ 0.23 \\

& PatchTST
& 10.38 $\pm$ 0.21
& \textbf{17.76 $\pm$ 0.13}
& \textbf{30.00 $\pm$ 0.08}
& 10.54 $\pm$ 0.24
& 17.87 $\pm$ 0.27
& 30.02 $\pm$ 0.21
& 10.52 $\pm$ 0.22
& 18.07 $\pm$ 0.21
& 29.57 $\pm$ 0.19 \\

& Chronos-bolt-tiny$^{\text{ZS}}$
& 12.21 & 20.55 & 34.24
& 11.77 & 19.89 & 33.01
& 11.29 & 19.08 & 31.26 \\

& Chronos-bolt-tiny$^{\text{FT}}$
& 10.11 $\pm$ 0.06
& 17.89 $\pm$ 0.06
& 31.06 $\pm$ 0.14
& 10.14 $\pm$ 0.08
& \textbf{17.54 $\pm$ 0.05}
& \textbf{29.67 $\pm$ 0.09}
& \textbf{9.95 $\pm$ 0.12}
& 17.36 $\pm$ 0.08
& \textbf{29.12 $\pm$ 0.07} \\

& Chronos-bolt-mini$^{\text{ZS}}$
& 12.12 & 20.71 & 34.62
& 11.62 & 19.85 & 33.00
& 11.26 & 18.92 & 31.34 \\

& Chronos-bolt-mini$^{\text{FT}}$
& \textbf{10.07 $\pm$ 0.05}
& 17.83 $\pm$ 0.05
& 31.05 $\pm$ 0.17
& \textbf{10.05 $\pm$ 0.08}
& 17.56 $\pm$ 0.19
& 29.83 $\pm$ 0.22
& 9.98 $\pm$ 0.07
& \textbf{17.27 $\pm$ 0.07}
& 29.30 $\pm$ 0.15 \\

& Chronos2-small$^{\text{ZS}}$
& 11.38 & 19.79 & 34.14
& 11.19 & 18.81 & 31.54
& 10.96 & 18.38 & 30.82 \\

& Chronos2-small$^{\text{FT}}$
& 10.45 $\pm$ 0.11
& 18.43 $\pm$ 0.19
& 31.81 $\pm$ 0.40
& 10.41 $\pm$ 0.16
& 17.94 $\pm$ 0.11
& 30.35 $\pm$ 0.31
& 10.50 $\pm$ 0.11
& 17.77 $\pm$ 0.08
& 29.50 $\pm$ 0.23 \\

& TimesFM-2.5$^{\text{ZS}}$
& 10.92 & 19.75 & 33.67
& 10.33 & 18.29 & 31.06
& 10.03 & 17.57 & 29.77 \\

& TimesFM-2.5$^{\text{FT}}$
& 10.57 $\pm$ 0.27
& 19.52 $\pm$ 0.30
& 32.68 $\pm$ 0.37
& 10.26 $\pm$ 0.09
& 17.86 $\pm$ 0.05
& 30.77 $\pm$ 0.10
& 9.96 $\pm$ 0.04
& 17.46 $\pm$ 0.05
& 29.64 $\pm$ 0.05 \\

\midrule

% ================= Non-diabetes and T2D (RMSE) =================
\multirow{13}{*}[-10pt]{%
  \rotatebox{90}{\textbf{Non-diabetes \& T2D}}%
}
& LOCF
& 8.10 & 11.53 & 15.74
& 8.10 & 11.53 & 15.74
& 8.10 & 11.53 & 15.74 \\

& AutoARIMA
& 8.31 & 13.33 & 20.28
& 7.74 & 11.61 & 15.95
& 7.50 & 11.02 & 15.22 \\

& Elastic Net
& 7.48 & 10.83 & 14.59
& 7.35 & 10.51 & 14.08
& 7.33 & 10.37 & 13.57 \\

& LSTM
& 7.64 $\pm$ 0.05
& 10.82 $\pm$ 0.08
& 14.47 $\pm$ 0.07
& 7.48 $\pm$ 0.07
& 10.49 $\pm$ 0.06
& 14.09 $\pm$ 0.07
& 7.49 $\pm$ 0.06
& 10.45 $\pm$ 0.08
& 13.54 $\pm$ 0.11 \\

& PatchTST
& \textbf{7.38 $\pm$ 0.06}
& 10.51 $\pm$ 0.06
& \textbf{13.79 $\pm$ 0.07}
& 7.40 $\pm$ 0.09
& 10.51 $\pm$ 0.09
& 13.62 $\pm$ 0.10
& 7.47 $\pm$ 0.12
& 10.33 $\pm$ 0.05
& 13.31 $\pm$ 0.10 \\

& Chronos-bolt-tiny$^{\text{ZS}}$
& 8.56 & 12.26 & 16.48
& 8.26 & 11.77 & 15.99
& 7.83 & 10.95 & 14.53 \\

& Chronos-bolt-tiny$^{\text{FT}}$
& 7.40 $\pm$ 0.12
& \textbf{10.34 $\pm$ 0.06}
& 13.85 $\pm$ 0.05
& 7.25 $\pm$ 0.13
& \textbf{10.17 $\pm$ 0.05}
& \textbf{13.42 $\pm$ 0.02}
& 7.12 $\pm$ 0.06
& 10.01 $\pm$ 0.04
& 13.03 $\pm$ 0.05 \\

& Chronos-bolt-mini$^{\text{ZS}}$
& 8.60 & 12.34 & 16.91
& 8.27 & 11.78 & 16.21
& 7.86 & 10.96 & 14.71 \\

& Chronos-bolt-mini$^{\text{FT}}$
& \textbf{7.38 $\pm$ 0.15}
& 10.36 $\pm$ 0.06
& 13.84 $\pm$ 0.04
& \textbf{7.08 $\pm$ 0.05}
& 10.18 $\pm$ 0.06
& 13.44 $\pm$ 0.03
& \textbf{7.01 $\pm$ 0.04}
& \textbf{9.99 $\pm$ 0.03}
& \textbf{13.01 $\pm$ 0.05} \\

& Chronos2-small$^{\text{ZS}}$
& 7.96 & 11.28 & 15.29
& 7.71 & 10.58 & 14.11
& 7.46 & 10.18 & 13.30 \\

& Chronos2-small$^{\text{FT}}$
& 7.47 $\pm$ 0.02
& 10.51 $\pm$ 0.02
& 14.11 $\pm$ 0.04
& 7.40 $\pm$ 0.08
& 10.33 $\pm$ 0.04
& 13.85 $\pm$ 0.12
& 7.26 $\pm$ 0.09
& 10.00 $\pm$ 0.04
& 13.09 $\pm$ 0.10 \\

& TimesFM-2.5$^{\text{ZS}}$
& 7.82 & 11.18 & 15.12
& 7.32 & 10.54 & 14.03
& 7.13 & 10.14 & 13.41 \\

& TimesFM-2.5$^{\text{FT}}$
& 7.60 $\pm$ 0.11
& 11.11 $\pm$ 0.05
& 15.02 $\pm$ 0.22
& 7.24 $\pm$ 0.03
& 10.45 $\pm$ 0.03
& 14.20 $\pm$ 0.03
& 7.14 $\pm$ 0.03
& 10.21 $\pm$ 0.03
& 13.51 $\pm$ 0.06 \\

\bottomrule
\end{tabular}%
}
\endgroup

\begin{minipage}{0.95\linewidth}
\footnotesize
Results are reported as mean $\pm$ SD across repeated runs when applicable.
$^{\mathrm{ZS}}$: Zero-shot foundation models (deterministic);
$^{\mathrm{FT}}$: Fine-tuned foundation models.
\end{minipage}

\end{table}

\subsubsection{Fine-tuning enables more consistent gains for selected foundation models}

In contrast to zero-shot deployment, fine-tuning leads to substantial and consistent performance improvements for several foundation models (Table~\ref{tab:ood_gain_pct_combined}). The most pronounced gains are observed for the Chronos-Bolt variants. On the T1D cohort, fine-tuning reduces RMSE by 6.5--18.4\% for Chronos-Bolt-Tiny and Mini across all evaluated context--horizon combinations. Similar trends are observed in the Non-diabetes/T2D cohort, where RMSE reductions typically range from 8.6--18.2\%. These gains frequently shift foundation models from underperforming strong baselines in zero-shot mode to becoming top-performing methods after fine-tuning across prediction horizons.

In contrast, the benefits of fine-tuning are more modest for larger foundation models. Chronos2-Small exhibits consistent but smaller improvements, generally below 8\% across settings. In several Non-diabetes/T2D configurations, TimesFM-2.5 shows slight performance degradation after fine-tuning, particularly for longer prediction horizons. These patterns may reflect diminishing returns from adaptation when model capacity is large relative to the available training data.

Table~\ref{tab:ood_gain_pct_combined} shows that fine-tuning is important for realizing strong CGM performance from pretrained foundation models, especially for smaller- and medium-capacity architectures, but does not yield uniform gains across all models and settings.

\begin{table}[htbp]
\centering
\renewcommand{\arraystretch}{1}
\caption{Fine-tuning improvement (\%) in RMSE across ID and OOD settings on the T1D and Non-diabetes/T2D cohorts.}
\label{tab:ood_gain_pct_combined}
\scalebox{0.7}{
\begin{tabular}{@{}p{5mm}@{} l cccccccc}
\toprule
                  &
\textbf{Ctx-Pred} &
\multicolumn{2}{c}{\textbf{Chronos-bolt-tiny}} &
\multicolumn{2}{c}{\textbf{Chronos-bolt-mini}} &
\multicolumn{2}{c}{\textbf{Chronos2\_small}} &
\multicolumn{2}{c}{\textbf{TimesFM-2.5}} \\
\cmidrule(lr){3-4} \cmidrule(lr){5-6} \cmidrule(lr){7-8} \cmidrule(lr){9-10}
& &
\textbf{ID} & \textbf{OOD} &
\textbf{ID} & \textbf{OOD} &
\textbf{ID} & \textbf{OOD} &
\textbf{ID} & \textbf{OOD} \\
\midrule

\multirow{9}{*}{\rotatebox[origin=c]{90}{\textbf{T1D}}}
& Ctx: 4h, Pred: 30m & 17.2 & 18.4 & 16.9 & 17.1 & 8.1 & 8.1 & 3.2 & 3.2 \\
& Ctx: 4h, Pred: 1h  & 12.9 & 12.9 & 13.9 & 12.9 & 6.9 & 7.2 & 1.2 & 2.0 \\
& Ctx: 4h, Pred: 2h  & 9.3  & 10.1 & 10.3 & 11.4 & 6.8 & 7.6 & 2.9 & 4.2 \\
\cmidrule(l){2-10}
& Ctx: 12h, Pred: 30m & 13.8 & 14.9 & 13.5 & 14.8 & 7.0 & 6.1 & 0.7 & 0.8 \\
& Ctx: 12h, Pred: 1h  & 11.8 & 11.9 & 11.5 & 11.1 & 4.7 & 4.8 & 2.4 & 2.2 \\
& Ctx: 12h, Pred: 2h  & 10.1 & 10.7 & 9.6  & 10.2 & 3.8 & 3.8 & 0.9 & 1.1 \\
\cmidrule(l){2-10}
& Ctx: 24h, Pred: 30m & 11.9 & 12.7 & 11.4 & 12.0 & 4.2 & 4.7 & 0.6 & 0.6 \\
& Ctx: 24h, Pred: 1h  & 9.0  & 8.5  & 8.7  & 8.6  & 3.3 & 3.8 & 0.6 & 0.1 \\
& Ctx: 24h, Pred: 2h  & 6.9  & 7.2  & 6.5  & 6.5  & 4.3 & 4.7 & 0.4 & 0.6 \\
\midrule

\multirow{9}{*}{\rotatebox[origin=c]{90}{\textbf{Non-diabetes \& T2D}}}
& Ctx: 4h, Pred: 30m & 13.6 & 12.7 & 14.2 & 13.3 & 6.2 & 6.0 & 2.9 & 1.2 \\
& Ctx: 4h, Pred: 1h  & 15.6 & 15.8 & 16.1 & 16.9 & 6.8 & 7.8 & 0.6 & 3.2 \\
& Ctx: 4h, Pred: 2h  & 16.0 & 16.0 & 18.2 & 17.8 & 7.7 & 6.9 & 0.7 & 1.1 \\
\cmidrule(l){2-10}
& Ctx: 12h, Pred: 30m & 12.3 & 12.8 & 14.4 & 14.4 & 4.0 & 4.3 & 1.1 & 0.7 \\
& Ctx: 12h, Pred: 1h  & 13.6 & 14.3 & 13.6 & 14.3 & 2.3 & 3.3 & -0.9 & -0.1 \\
& Ctx: 12h, Pred: 2h  & 16.1 & 16.8 & 17.1 & 17.7 & 1.9 & 2.3 & -1.2 & -0.8 \\
\cmidrule(l){2-10}
& Ctx: 24h, Pred: 30m & 9.1  & 9.3  & 10.9 & 10.5 & 2.6 & 3.8 & -0.1 & -0.1 \\
& Ctx: 24h, Pred: 1h  & 8.6  & 9.0  & 8.8  & 8.8  & 1.7 & 3.1 & -0.7 & -1.2 \\
& Ctx: 24h, Pred: 2h  & 10.3 & 11.8 & 11.5 & 12.6 & 1.6 & 2.7 & -0.7 & -0.5 \\
\bottomrule
\end{tabular}
}
\end{table}

\subsubsection{Fine-tuning yields comparable gains on ID and OOD}

Table~\ref{tab:ood_gain_pct_combined} shows that fine-tuning yields comparable relative RMSE improvements on both ID and OOD test sets across most context--horizon configurations. Across both cohorts and all evaluated models, the difference between ID and OOD improvement percentages is typically small (generally within 1 percentage point), indicating that the gains from fine-tuning transfer consistently to unseen subjects.

This consistency suggests that lightweight fine-tuning primarily improves alignment to CGM-specific temporal dynamics rather than inducing overfitting to in-distribution data, and that the resulting performance gains remain robust under distribution shift.

\subsection{Multimodal Integration for Glucose Forecasting}

\subsubsection{Ablation study of multimodal integration}

Table~\ref{tab:ablation_multimodal} summarizes ablation results assessing the contributions of individual modalities and image supervision strategies under a fixed 24-hour context and a 2-hour prediction horizon. Relative to the CGM-only baseline, incorporating meal-related information consistently improves forecasting performance.

Using food image information alone yields modest but consistent gains. Nutrition-aligned image embeddings outperform category-supervised embeddings in overall RMSE, indicating that visually inferred nutritional properties are more informative for glucose forecasting than food category recognition alone. In contrast, incorporating macronutrient features produces larger improvements, underscoring the importance of explicit nutritional information, particularly for modeling postprandial glucose dynamics.

Combining food image representations with macronutrients features yields additional gains beyond either modality alone, reflecting complementary contributions from visual and nutritional information. Among all configurations, jointly integrating nutrition-aligned image embeddings and macronutrients achieves the lowest overall RMSE. While the reduction in overall RMSE is modest (approximately 3\% relative to the CGM-only baseline), postprandial RMSE is reduced by around 15\%, highlighting substantial benefits in meal-related forecasting windows.

\begin{table}[htbp]
\centering

\caption{Ablation study of multimodal integration and image supervision strategy for CGM forecasting.}
\label{tab:ablation_multimodal}

\begin{minipage}{0.82\linewidth}
\centering
\renewcommand{\arraystretch}{0.9}
\small

\begin{tabularx}{\linewidth}{@{}Xcc@{}}
\toprule
Model configuration &
Overall RMSE &
Postprandial RMSE \\
\midrule

CGM only (Chronos)
& 16.90 (0.02)
& 28.79 (0.26) \\

- Image (Cat)
& 16.72 (0.08)
& 25.22 (0.41) \\

- Image (Nutr)
& 16.64 (0.06)
& 25.17 (0.17) \\

- Nutri
& 16.56 (0.16)
& \textbf{24.17 (0.52)} \\

- Image (Cat) + Nutri
& 16.49 (0.13)
& 24.41 (0.75) \\

- Image (Nutr) + Nutri
& \textbf{16.39 (0.07)}
& 24.51 (0.55) \\

\bottomrule
\end{tabularx}

\par\medskip

{\footnotesize
\raggedright
\noindent
RMSE values are reported as mean (standard deviation) for a
24-hour context and a 2-hour prediction horizon, averaged over 10 runs.
Postprandial RMSE is evaluated within 0--2 hours after meal events.
Parenthetical terms in the model configuration column denote image
supervision strategies. Cat denotes category-supervised image embeddings;
Nutr denotes nutrition-aligned image embeddings; Nutri denotes
macronutrient features.
\par
}

\end{minipage}
\end{table}

Table~\ref{tab:alt_encoders} reports results using alternative CGM backbones. CatBoost and LSTM showed a similar pattern: adding meal-related modalities generally improved performance, especially postprandial RMSE. Compared with these alternatives, the Chronos-based framework achieved lower overall RMSE.

\begin{table}[htbp]
\centering

\caption{Alternative CGM encoders for multimodal CGM forecasting (CatBoost vs. LSTM).}
\label{tab:alt_encoders}

\begin{minipage}{0.82\linewidth}
\centering
\renewcommand{\arraystretch}{0.5}
\small

\begin{tabularx}{\linewidth}{Xcc}
\toprule
Model configuration &
Overall RMSE &
Postprandial RMSE \\
\midrule

CGM only (CatBoost)
& 17.78 (0.01)
& 27.48 (4.37) \\

- Image (Cat)
& 17.24 (0.02)
& 24.03 (2.38) \\

- Image (Nutr)
& 17.26 (0.02)
& 24.22 (2.32) \\

- Nutri
& \textbf{16.91 (0.01)}
& \textbf{23.49 (2.18)} \\

- Image (Cat) + Nutri
& 17.03 (0.03)
& 23.80 (2.43) \\

- Image (Nutr) + Nutri
& 17.03 (0.02)
& 23.91 (2.45) \\

\midrule

CGM only (LSTM)
& 17.85 (0.10)
& 27.16 (0.72) \\

- Image (Cat)
& 17.68 (0.07)
& 24.81 (0.09) \\

- Image (Nutr)
& 17.56 (0.10)
& 24.63 (0.41) \\

- Nutri
& 17.57 (0.14)
& 25.88 (0.76) \\

- Image (Cat) + Nutri
& 17.49 (0.02)
& \textbf{24.56 (0.13)} \\

- Image (Nutr) + Nutri
& \textbf{17.40 (0.11)}
& 24.88 (0.61) \\

\bottomrule
\end{tabularx}

\par\medskip

{\footnotesize
\raggedright
\noindent
CatBoost is trained on raw CGM readings and serves as a
reference baseline, whereas LSTM represents an alternative CGM sequence
encoder within the same multimodal framework. All other settings follow
Table~\ref{tab:ablation_multimodal}.
\par
}

\end{minipage}
\end{table}

Beyond RMSE-based forecasting accuracy, we further evaluated whether the models captured the magnitude of meal-induced glucose excursions. Table~\ref{tab:pgi} reports the correlation between predicted and observed PGI, providing an event-level assessment of postprandial excursion characterization rather than pointwise prediction error alone.

Across all three forecasting backbones, adding meal-related information improves PGI correlation relative to the CGM-only baseline. For Chronos, the correlation increases from 0.68 with CGM only to 0.76 with macronutrients, while the full Image+Nutrition model achieves 0.74. Similar trends are observed for LSTM and CatBoost, indicating that nutritional context provides robust information for estimating postprandial glucose excursions. Although Chronos does not always achieve the lowest postprandial RMSE, it yields substantially higher PGI correlations than LSTM and CatBoost across all modality configurations. This suggests that Chronos-based representations better preserve the relative magnitude of meal-induced glucose excursions. In contrast, the lower postprandial RMSE achieved by some alternative models may reflect more conservative or smoothed pointwise predictions, which reduce local errors but compress the variability of postprandial peaks.

\begin{table}[htbp]
\centering

\caption{Correlation between predicted and true postprandial glucose increment.}
\label{tab:pgi}

\begin{minipage}{0.75\linewidth}
\centering
\renewcommand{\arraystretch}{0.6}
\setlength{\tabcolsep}{3pt}
\small

\begin{tabularx}{\linewidth}{Xccc}
\toprule
Model configuration &
Chronos &
LSTM &
CatBoost \\
\midrule

CGM only
& 0.68 (0.005)
& 0.55 (0.008)
& 0.48 (0.005) \\

- Image
& 0.73 (0.007)
& 0.62 (0.010)
& 0.57 (0.009) \\

- Nutri
& 0.76 (0.007)
& 0.67 (0.005)
& 0.58 (0.002) \\

- Image + Nutri
& 0.74 (0.006)
& 0.62 (0.007)
& 0.59 (0.005) \\

\bottomrule
\end{tabularx}

\par\medskip

{\footnotesize
\raggedright
\noindent
Values are reported as mean correlation with standard
errors in parentheses. PGI was defined as the increase from pre-meal
glucose to the postprandial peak within 2 hours. Image denotes
nutrition-aligned food image embeddings.
\par
}

\end{minipage}
\end{table}

\begin{figure}[htbp]
  \centering
  \includegraphics[width=0.8\linewidth]{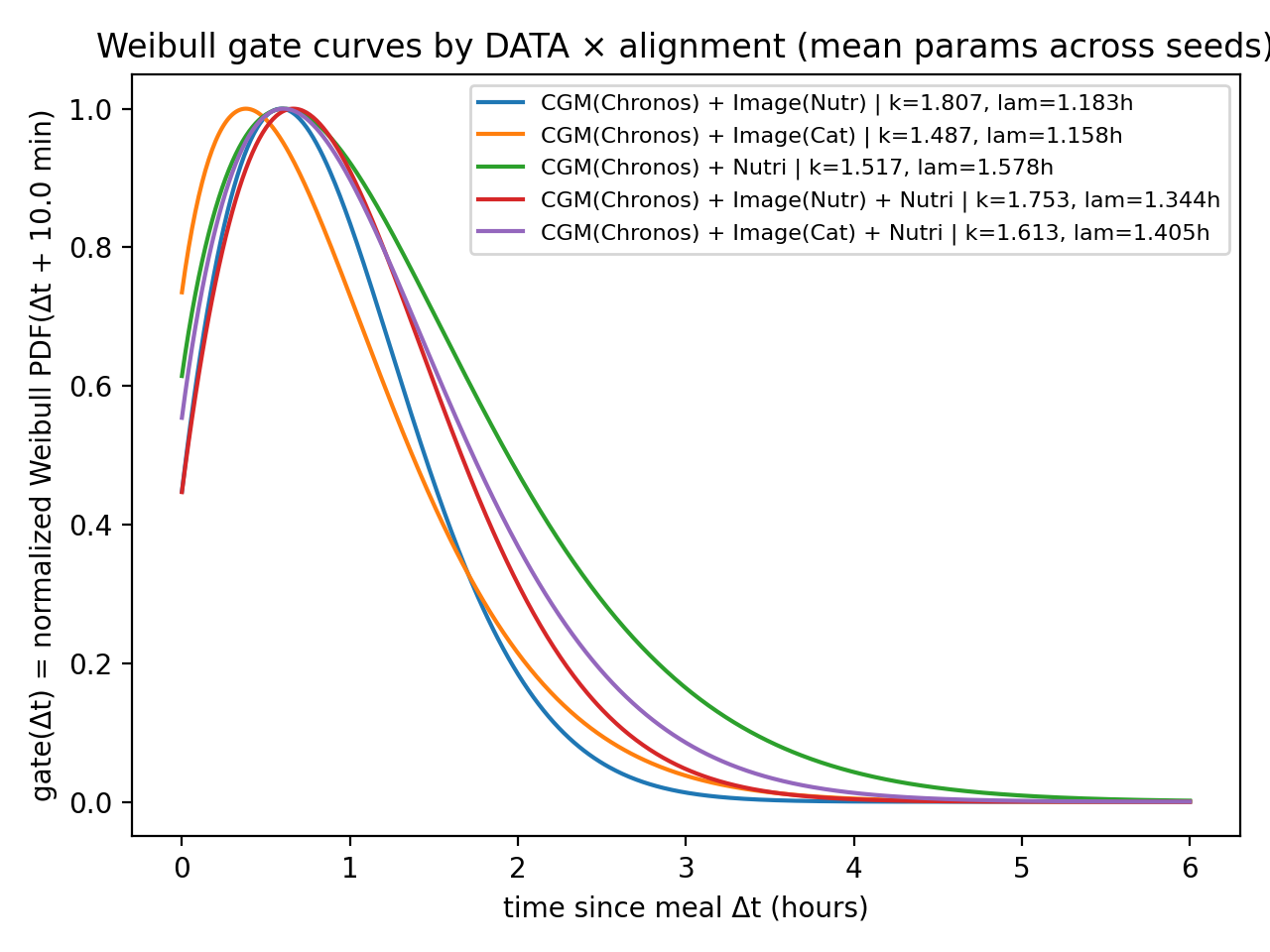}
  \caption{Learned Weibull-based temporal gating functions.}
  \label{fig:gate}
\end{figure}

This pronounced postprandial improvement is further supported by the learned Weibull-based temporal gating functions (Figure~\ref{fig:gate}), which consistently peak approximately 40--60 minutes after meal events and gradually diminish toward zero by around 3.5 hours post-meal. This behavior aligns with the typical timing of early postprandial glucose responses and confirms that the residual correction branch is adaptively emphasized during meal-proximal periods while being automatically attenuated at longer delays.

\begin{figure}[htbp]
  \centering
  \includegraphics[width=\textwidth]{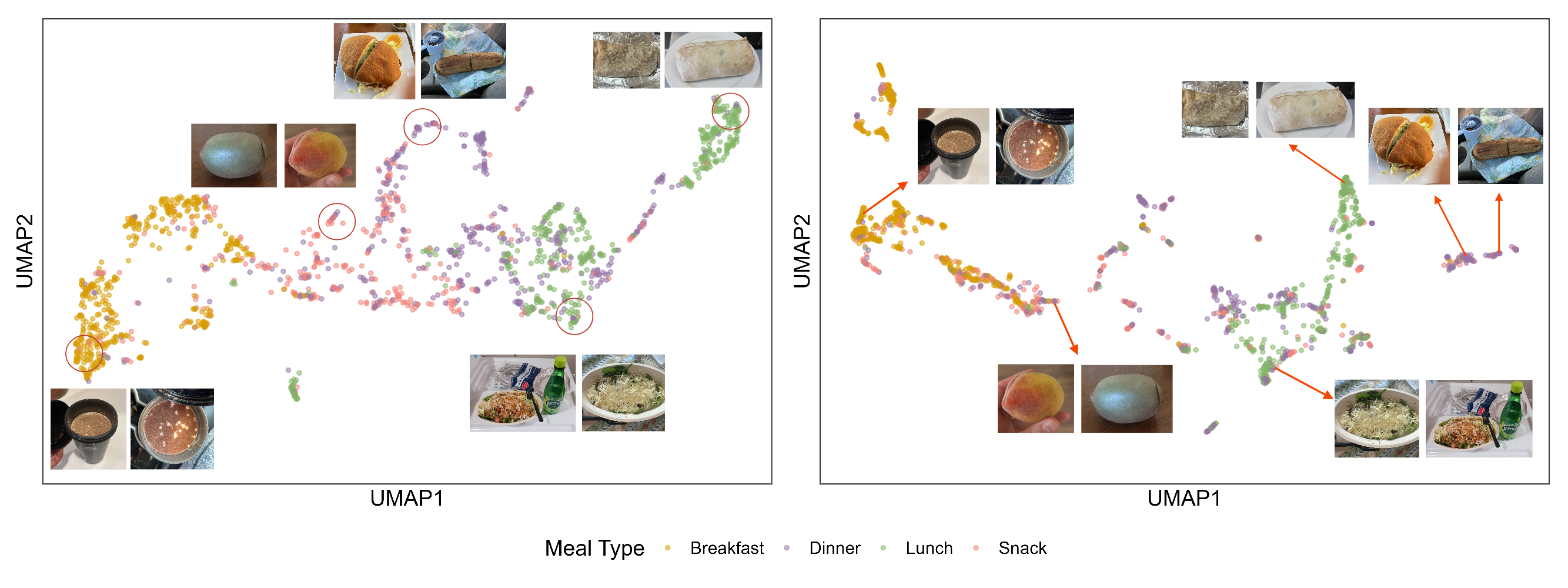}
  \caption{Visualization of food image embeddings under different supervision strategies. UMAP projections of food image embeddings extracted using two supervision schemes. Left: category-supervised image embeddings, trained to discriminate high-level meal categories. Right: nutrition-aligned image embeddings, trained with nutritional supervision. Representative food images are overlaid to illustrate local neighborhood semantics across the embedding space.}
  \label{fig:embed_position}
\end{figure}

\subsubsection{Analysis of nutrition-aligned visual representations}

To qualitatively examine how the supervision objective affects food image representations, we visualized UMAP projections of embeddings extracted from the same visual backbone trained with either category-level or nutrition-aligned supervision (Figure~\ref{fig:embed_position}). Under category supervision (left), embeddings were largely organized by visual appearance, so visually similar foods could appear nearby even when their nutritional profiles differed. In contrast, nutrition-aligned supervision (right) produced an embedding space with somewhat clearer separation according to macronutrient profiles.

These patterns suggest that nutrition-aligned supervision can shift visual representations toward nutritionally relevant structure, making the embeddings more aligned with factors related to physiological glucose responses. This may help explain the modest improvement over category-supervised embeddings in multimodal CGM forecasting.

\section{Limitations}

This study has several limitations related to data availability, multimodal measurement, and model design.

First, benchmarking time-series foundation models is constrained by the scale and characteristics of currently available public CGM datasets. Although multiple cohorts were evaluated under a unified protocol, existing datasets remain limited in subject count and temporal coverage relative to those used for foundation model pretraining. In addition, the use of fixed, clinically motivated historical context lengths may limit the ability of foundation models to exploit longer-range temporal dependencies. Thus, observed performance differences should be interpreted in light of these data and context-length constraints.

Second, the multimodal analysis was limited to the CGMacros dataset. Although CGMacros uniquely provides temporally aligned CGM signals, food images, and macronutrient information, the cohort is modest in size, and its dietary patterns and participant characteristics may not fully represent broader CGM-using populations. Dietary records may also contain occasional inaccuracies in macronutrient annotation, introducing noise that could attenuate estimated nutritional effects.

Third, our framework emphasized controlled comparisons across input modalities and supervision strategies rather than exhaustive optimization of multimodal fusion architectures. While this design supports interpretable ablation, it may not capture more complex cross-modal interactions. Future work should explore more expressive fusion mechanisms and validate multimodal CGM forecasting models in larger, more diverse, and externally collected cohorts.

\section{Conclusion}
This work presents a systematic evaluation of time-series foundation models and multimodal dietary information for CGM forecasting under unified and clinically relevant settings. Our results show that off-the-shelf foundation models do not consistently outperform strong task-specific baselines, whereas lightweight fine-tuning is often critical for achieving reliable performance gains across forecasting horizons and populations. We further show that incorporating dietary context improves postprandial glucose prediction, with explicit macronutrient features providing the largest gains and food images offering complementary benefits when trained with nutrition-aligned supervision. Overall, these findings suggest that foundation models can be effective for CGM forecasting when appropriately adapted, and that task-aligned multimodal dietary information provides useful signals beyond CGM alone.

\section*{Acknowledgment}

Generative AI tools were used in a limited capacity to assist with language editing and refinement of the manuscript. All algorithms, experimental protocols, and reported results were designed, verified, and validated by the authors.

\bibliographystyle{unsrtnat}
\bibliography{JHBI_IEEEtran}

\clearpage
\section*{Supplementary Material}
\addcontentsline{toc}{section}{Supplementary Material}

% Supplement-specific numbering: S1, S2, ...; Table S1; Figure S1; Equation S1.
\setcounter{section}{0}
\setcounter{subsection}{0}
\setcounter{table}{0}
\setcounter{figure}{0}
\setcounter{equation}{0}
\renewcommand{\thesection}{S\arabic{section}}
\renewcommand{\thesubsection}{S\arabic{section}.\arabic{subsection}}
\renewcommand{\thetable}{S\arabic{table}}
\renewcommand{\thefigure}{S\arabic{figure}}
\renewcommand{\theequation}{S\arabic{equation}}
% Distinct PDF destinations prevent duplicate-anchor warnings after counter resets.
\renewcommand{\theHsection}{supp.\arabic{section}}
\renewcommand{\theHsubsection}{supp.\arabic{section}.\arabic{subsection}}
\renewcommand{\theHtable}{supp.\arabic{table}}
\renewcommand{\theHfigure}{supp.\arabic{figure}}
\renewcommand{\theHequation}{supp.\arabic{equation}}

\section{Preprocessing Details}
\label{sec:preprocessing}

All CGM datasets were processed using a unified preprocessing pipeline designed to ensure consistency and reproducibility across studies. The pipeline aligned irregular CGM measurements to a uniform 5-minute grid, handled missing values using limited-gap interpolation, and constructed leakage-free data splits for forecasting evaluation. Final processed time series were exported in a standardized long-format representation with consistent subject-, segment-, and dataset-level identifiers.

When demographic information was available, analyses were restricted to adult participants (age $\ge$ 18 years). For studies involving closed-loop or automated insulin delivery systems, CGM data were restricted to periods reflecting natural glycemic dynamics. Specifically, CGM measurements prior to pump run-in or randomization were removed \cite{Anderson2016ClosedLoop,Lynch2022InsulinOnlyBP,Brown2019ClosedLoop}, and participants using hybrid closed-loop systems were excluded where applicable \cite{Brown2019ClosedLoop,Lynch2022InsulinOnlyBP}. CGM measurements overlapping documented adverse event windows were also excluded, with an additional one-day buffer following event resolution when reported.

CGM timestamps were rounded to the nearest 5-minute interval and mapped onto a complete subject-specific 5-minute grid. When multiple observations mapped to the same timestamp, their mean value was used. Missing glucose values were linearly interpolated only for short gaps of at most 24 consecutive 5-minute intervals (120 minutes). Longer gaps were not interpolated and were treated as discontinuities in the time series.

Following interpolation, contiguous runs of non-missing glucose values were identified as segments. Only segments containing at least 336 observations (28 hours) were retained. To mitigate boundary effects and early sensor instability, the first 24 observations (2 hours) of each segment were discarded, yielding a minimum effective segment length of 312 observations (26 hours).

To evaluate generalization to unseen individuals, a subject-level OOD test set was constructed independently within each dataset by randomly selecting 20\% of subjects. The remaining 80\% of subjects were split chronologically within subject into training, validation, and internal test sets using an approximately 2:1:1 ratio.

For multimodal forecasting experiments, the CGMacros dataset was processed using an additional modality-specific pipeline. Dietary records were filtered to retain only meal events with valid food images depicting the consumed items, excluding post-meal photographs. Meal timestamps were rounded to the nearest 5-minute interval to align with the CGM grid. Each retained meal was represented as a multimodal event consisting of a timestamp, structured nutrient information, and a corresponding food image embedding.

All multimodal data splits were performed strictly chronologically at the subject level. For each subject, the resampled 5-minute CGM time series was divided into training, validation, and test sets using a 6:2:2 ratio. Two subjects exhibiting prolonged recording gaps were segmented at gap boundaries, and one subject with insufficient or irregular data was excluded from analysis.

\section{Hyperparameters}

Hyperparameter optimization was conducted independently for each model and each context--horizon configuration. For non-deep-learning baselines, including LOCF, AutoARIMA, and Elastic Net, either no hyperparameter tuning or a simple grid search was used following standard practice. For all deep learning models, including LSTM, PatchTST, and fine-tuned foundation models, hyperparameters were selected via automated hyperparameter optimization, with validation MSE as the optimization objective.

PatchTST and Chronos models were trained using AutoGluon. Unless explicitly specified in Table~\ref{tab:hpo_space}, all remaining training configurations followed AutoGluon’s default settings. To ensure consistency across models, no automated ensembling, feature engineering, or model-specific heuristics provided by AutoGluon were enabled. Only the hyperparameters listed in Table~\ref{tab:hpo_space} were allowed to vary during optimization.

Fine-tuning strategies for foundation models were determined based on model scale and computational constraints. Chronos-Bolt models were fine-tuned by updating all model parameters. In contrast, Chronos2-Small and TimesFM~2.5 were adapted using Low-Rank Adaptation (LoRA), with pretrained backbone parameters frozen. The selected hyperparameter configurations for all models are reported in Tables~\ref{tab:T1Dhyper} and~\ref{tab:T2Dhyper}.

\section{Visual Encoder Fine-tuning}
\label{sec:CV_pretraining}

All candidate visual backbone models were fine-tuned end-to-end on the Food-101 dataset prior to downstream use. Based on preliminary model selection experiments, full end-to-end fine-tuning was used for all visual encoders in subsequent experiments. The selected visual backbone (ConvNeXT-base) was then applied to extract food image embeddings for multimodal glucose forecasting on the CGMacros dataset.

The visual encoder produced 1024-dimensional image embeddings. To obtain lower-dimensional representations suitable for multimodal integration, Uniform Manifold Approximation and Projection (UMAP) was applied as a post-processing step. UMAP dimensionality was selected based on trustworthiness diagnostics evaluated over a range of target embedding dimensions.

\begin{figure}[htbp]
  \centering
  \includegraphics[width=0.6\linewidth]{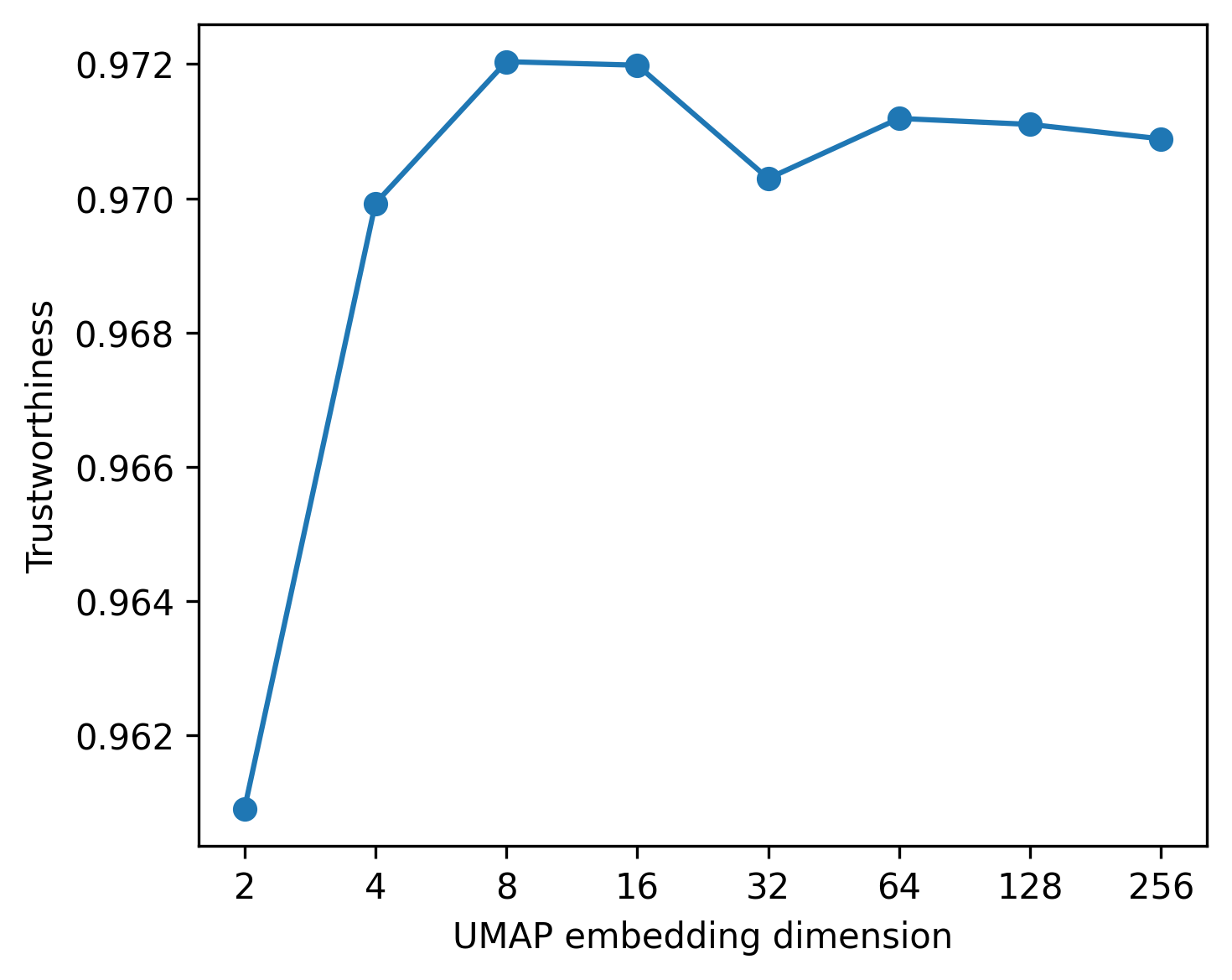}
  \caption{Trustworthiness of UMAP embeddings derived from category-supervised food image representations, evaluated as a function of embedding dimensionality.}
  \label{fig:trustworthiness}
\end{figure}

\begin{figure}[htbp]
  \centering
  \includegraphics[width=0.6\linewidth]{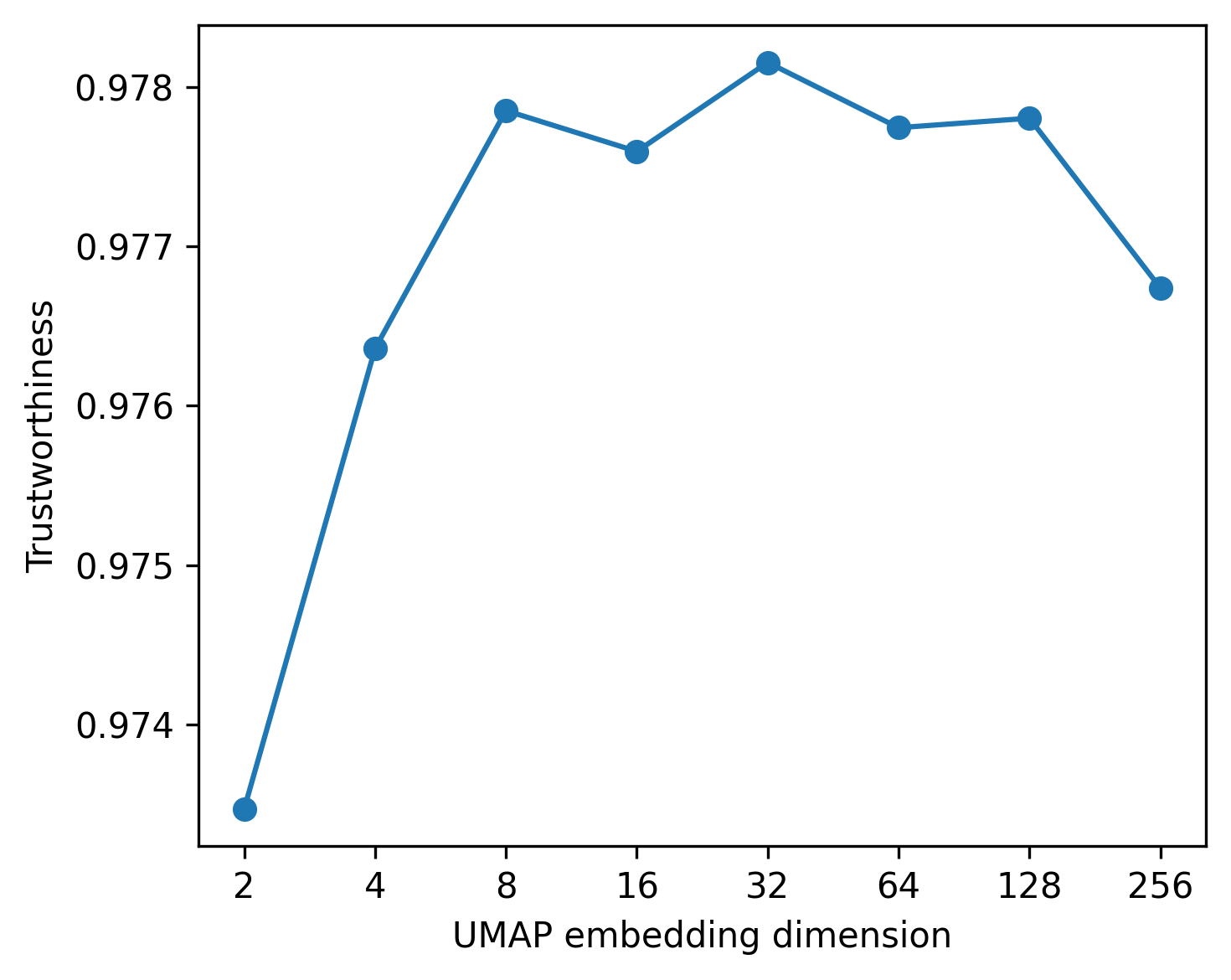}
  \caption{Trustworthiness of UMAP embeddings derived from nutrition-aligned food image representations, evaluated as a function of embedding dimensionality.}
  \label{fig:trustworthiness_aligned}
\end{figure}

Trustworthiness curves for UMAP embeddings derived from category-supervised and nutrition-aligned image representations are provided in Figures~\ref{fig:trustworthiness} and~\ref{fig:trustworthiness_aligned}. Based on these diagnostics, an embedding dimension of $d=8$ was selected and used consistently across all multimodal forecasting experiments.

\section{Computational Resources}

All experiments were conducted on Linux-based systems equipped with NVIDIA T4 GPUs. All deep learning models, including foundation models and multimodal architectures, were trained using a single-GPU configuration without distributed or multi-node parallelism. Hyperparameter optimization and repeated experimental runs were executed sequentially.

Visual encoder fine-tuning and multimodal forecasting experiments were performed on GPU resources, while non-neural baseline models were trained on CPU.

\section{Supporting Tables of Model Comparison for Reproducibility}

This subsection provides supporting reference tables corresponding to the experimental findings summarized in the main text. These tables are included solely to improve completeness, transparency, and reproducibility.

Tables~\ref{tab:cgm_perf_T1D} and~\ref{tab:cgm_perf_T2D} report the full numerical results for model performance stratified by cohort. In contrast to the main text, which presents RMSE results for conciseness, these tables additionally include MAE values for all evaluated models, historical context lengths, and prediction horizons under the standardized evaluation protocol described in Section~\ref{sec:preprocessing}.

Tables~\ref{tab:ood_gain_T1D} and~\ref{tab:ood_gain_T2D} provide complementary reference values for performance changes before and after fine-tuning of foundation models under both ID and OOD evaluation settings. While the main text reports relative improvements in percentage form for clarity, these tables further present the corresponding RMSE values to facilitate direct numerical comparison and reproducibility.

\section{Additional Analyses for Multimodal Generalization and Context-Dependent Modality Effects}
\label{sec:supp_multimodal_additional}

\subsection{Subject-level generalization under leave-one-subject-out evaluation}

A potential concern in multimodal CGM forecasting is that the observed improvement from dietary inputs may be partially driven by subject-specific identity effects rather than by physiologically meaningful meal-related information. To further assess subject-level generalization, we conducted additional leave-one-subject-out (LOSO) experiments using the fine-tuned Chronos backbone. In each LOSO split, the held-out subject was completely excluded from model training and used only for evaluation.

Table~\ref{tab:loso_multimodal} summarizes the LOSO results. Compared with the CGM-only Chronos backbone, all multimodal configurations reduced RMSE under this stricter subject-level evaluation. The CGM-only model achieved an RMSE of 17.5, whereas image-only, nutrition-only, and combined image--nutrition models achieved RMSE values between 16.9 and 17.0. These results indicate that the benefit of multimodal dietary information persists even when evaluation is performed on subjects unseen during training.

This evaluation is more stringent than the standard within-subject chronological split used in the primary multimodal experiments, because the test subject is not observed during training. In addition, the LOSO setup reduces the possibility that improvements arise from memorizing subject-specific glycemic patterns. The consistent improvement of multimodal models over the CGM-only backbone suggests that dietary information contributes predictive signals that generalize beyond individual identity effects.

We also note that the LOSO Chronos-based multimodal models remain competitive with, and in some cases stronger than, standard-split non-foundation baselines reported in the main manuscript. This further supports the robustness of the learned CGM and dietary representations under a more conservative evaluation setting.

\subsection{Context-dependent effectiveness of dietary modalities}

We further examined whether the relative usefulness of image and structured nutrition inputs depends on meal context. Table~\ref{tab:meal_type_multimodal} reports postprandial RMSE stratified by meal type for image-only, nutrition-only, and combined image--nutrition models.

The relative effectiveness of each modality varied across meal types. For breakfast, participants consumed a standardized protein shake with nearly identical visual appearance. As a result, image features were less informative, and structured nutritional inputs provided stronger predictive value. For lunch, meals were consistently sourced from Chipotle, and nutrition labels could be reliably derived from restaurant menu information. In this setting, image and nutrition modalities were well aligned, and the combined model achieved the lowest RMSE. For dinner, nutritional information was self-reported and contained noticeable inaccuracies in some records. Consequently, the nutrition-only model underperformed the image-only model, and adding nutrition to image features degraded performance relative to using image features alone.

These results suggest that the value of each dietary modality is highly context-dependent and is strongly influenced by modality reliability and data quality. In particular, structured nutrition inputs are most useful when they are accurately recorded and well aligned with the consumed meal, whereas image features may be more robust when nutrition annotations are noisy or incomplete. This finding highlights an important practical consideration for multimodal glucose forecasting: future models may benefit from adaptive modality weighting or context-aware modality selection, especially when different input modalities vary in reliability across meal types or recording conditions.

\section{A Qualitative Nearest-Neighbor Example of Visual Embeddings}

\begin{figure}[htbp]
  \centering
  \includegraphics[width=\textwidth]{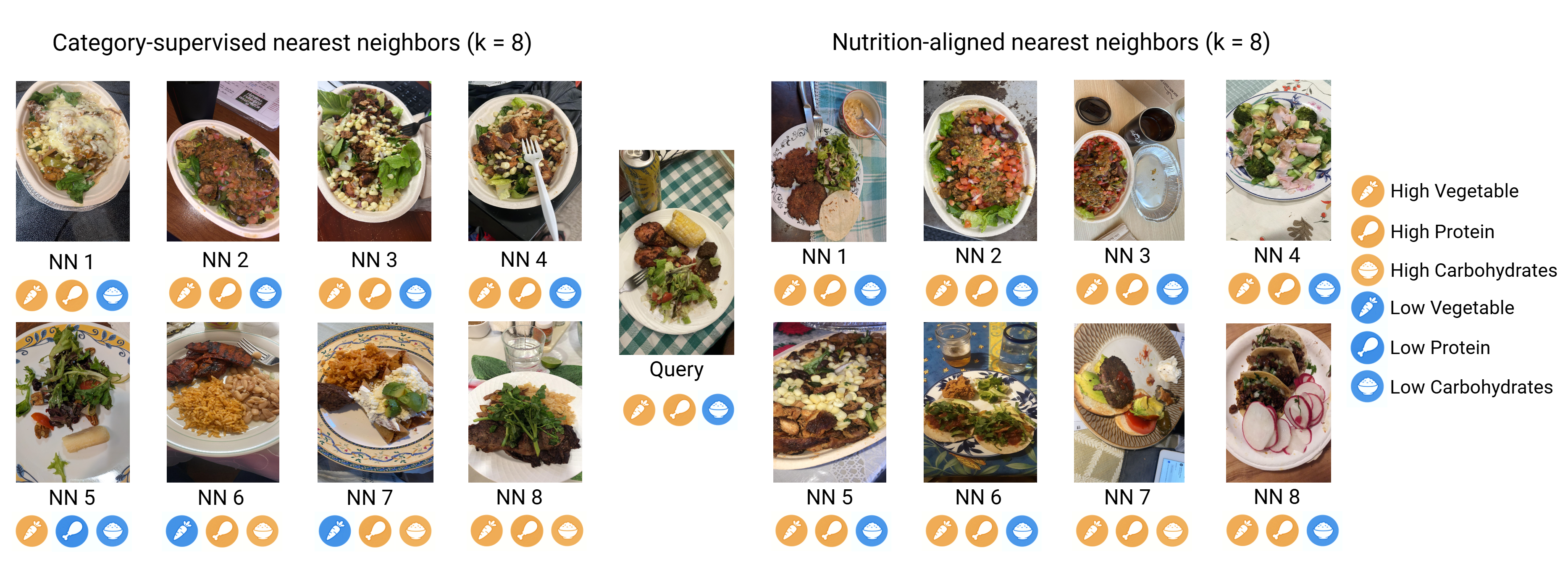}
  \caption{Nearest-neighbor examples for food image embeddings learned with different supervision strategies. For a given query image (center), the top $k=8$ nearest neighbors are shown for category-supervised (left) and nutrition-aligned (right) models. Icons denote relative vegetable (fiber), protein, and carbohydrate levels (high vs.\ low) associated with each meal.}
  \label{Fig:KNN}
\end{figure}

Figure~\ref{Fig:KNN} provides a qualitative illustration of the visual embedding spaces induced by category-supervised and nutrition-aligned training objectives using a nearest-neighbor retrieval example. Given a randomly selected food image from the CGMacros dataset, the top-$k$ nearest neighbors ($k=8$) are retrieved under each embedding space.

For each retrieved set, the associated meal records are shown to illustrate the corresponding dietary compositions. Under category-supervised embeddings, nearest neighbors tend to be visually similar, while exhibiting substantial variability in recorded nutritional attributes. Under nutrition-aligned embeddings, retrieved neighbors exhibit greater consistency in recorded vegetable, protein, and carbohydrate composition, while remaining visually similar.

This qualitative example is included as a supplementary diagnostic to document differences in the structure of the learned embedding spaces under the two training objectives.

\clearpage

\begin{table}
\centering
\caption{Hyperparameter search space for all evaluated models.}
\label{tab:hpo_space}
\renewcommand{\arraystretch}{1.25}

\begin{threeparttable}
\resizebox{\textwidth}{!}{%
\begin{tabular}{l c l l}
\toprule
Model & \# Parameters & Hyperparameter & Search range \\
\midrule
\multirow{5}{*}{LSTM}
  & \multirow{5}{*}{Varied}
  & lr           & $[10^{-4},\,3\times10^{-3}]$ \\
  &  & batch\_size   & $\{64,\,128,\,256\}$ \\
  &  & hidden\_dim   & $\{64,\,128,\,256\}$ \\
  &  & num\_layers   & $\{1,\,2,\,3\}$ \\
  &  & dropout       & $[0.1,\,0.3]$ \\
\midrule
\multirow{2}{*}{PatchTST}
  & \multirow{2}{*}{$\sim$25K}
  & lr           & $[10^{-4},\,3\times10^{-3}]$ \\
  &  & batch\_size   & $\{64,\,128,\,256\}$ \\
\midrule
\multirow{2}{*}{Chronos-bolt-tiny}
  & \multirow{2}{*}{9M}
  & lr           & $[10^{-5},\,10^{-3}]$ \\
  &  & batch\_size   & $\{64,\,128,\,256\}$ \\
\midrule
\multirow{2}{*}{Chronos-bolt-mini}
  & \multirow{2}{*}{21M}
  & lr           & $[10^{-5},\,10^{-3}]$ \\
  &  & batch\_size   & $\{64,\,128,\,256\}$ \\
\midrule
\multirow{2}{*}{Chronos2-small}
  & \multirow{2}{*}{28M}
  & lr           & $[10^{-5},\,10^{-3}]$ \\
  &  & batch\_size   & $\{64,\,128\}$ \\
\midrule
\multirow{4}{*}{TimesFM-2.5}
  & \multirow{4}{*}{200M}
  & lr                   & $[10^{-5},\,10^{-3}]$ \\
  &  & batch\_size           & $\{64,\,128,\,256\}$ \\
  &  & LoRA $(r,\alpha)$     & $\{(8,16),\,(16,32)\}$ \\
  &  & LoRA target modules   & $\{\texttt{out},\,(\texttt{out},\texttt{qkv\_proj}),$ \\
  &  &                       & $(\texttt{out},\texttt{qkv\_proj},\texttt{ff0},\texttt{ff1})\}$ \\
\bottomrule
\end{tabular}
}%

\begin{tablenotes}[flushleft]

\item Learning rate (lr) was tuned on a logarithmic scale within the specified range. Batch size was constrained by GPU memory for large foundation models. For Chronos2-small and TimesFM-2.5, low-rank adaptation (LoRA) was applied to selected modules only, while the backbone parameters were frozen. All hyperparameters were optimized using Bayesian optimization on the validation set.
\end{tablenotes}

\end{threeparttable}
\end{table}

\begin{table}
\caption{Hyperparameter configurations for all models in the T1D cohort, obtained via hyperparameter optimization (HPO).}
\label{tab:T1Dhyper}
\renewcommand{\arraystretch}{1.8}
\centering
\begin{threeparttable}
\resizebox{\textwidth}{!}{%
\begin{tabular}{llllllclllll}
\toprule
 & Hyperparams &  & \multicolumn{3}{c}{Ctx: 4-hr} & \multicolumn{3}{c}{Ctx: 12-hr} & \multicolumn{3}{c}{Ctx: 24-hr} \\
\cmidrule(lr){4-6} \cmidrule(lr){7-9} \cmidrule(lr){10-12}
 &  & Pred
 & 30-min & 1-hr & 2-hr
 & 30-min & 1-hr & 2-hr
 & 30-min & 1-hr & 2-hr \\
\midrule
\multirow{5}{*}{\rotatebox{90}{LSTM}}
 & lr          &  & 3.4e-4 & 3.4e-4 & 3.4e-4 & 3.4e-4 & 3.4e-4 & 3.4e-4 & 3.4e-4 & 3.4e-4 & 3.4e-4 \\
 & batch\_size &  & 64     & 64     & 64     & 64     & 64     & 64     & 64     & 64     & 64     \\
 & hidden\_dim &  & 128    & 128    & 128    & 128    & 128    & 128    & 128    & 128    & 128    \\
 & num\_layers &  & 2      & 2      & 2      & 2      & 2      & 2      & 2      & 2      & 2      \\
 & dropout     &  & 0.212  & 0.212  & 0.212  & 0.212  & 0.212  & 0.212  & 0.212  & 0.212  & 0.212  \\
\midrule
\multirow{2}{*}{\rotatebox{90}{PatchTST}}
 & lr          &  & 1.6e-3 & 1.6e-3 & 7.5e-4 & 1.6e-3 & 1.6e-3 & 7.5e-4 & 1.6e-3 & 7.5e-4 & 7.8e-4 \\
 & batch\_size &  & 256    & 256    & 64     & 256    & 256    & 64     & 256    & 64     & 64     \\
\midrule
\multirow{2}{*}{\rotatebox{90}{\shortstack{Chronos\\(bolt-tiny)}}}
 & lr          &  & 1.4e-4 & 4.2e-4 & 1.5e-4 & 4.7e-4 & 1.4e-4 & 7.5e-5 & 1.4e-4 & 7.5e-5 & 1.4e-4 \\
 & batch\_size &  & 256    & 256    & 64     & 128    & 256    & 256    & 256    & 256    & 256    \\
\midrule
\multirow{2}{*}{\rotatebox{90}{\shortstack{Chronos\\(bolt-mini)}}}
 & lr          &  & 7.5e-5 & 4.2e-4 & 1.5e-4 & 1.8e-4 & 1.4e-4 & 1.4e-4 & 5.8e-5 & 7.5e-5 & 4.2e-4 \\
 & batch\_size &  & 256    & 256    & 64     & 128    & 256    & 256    & 64     & 256    & 256    \\
\midrule
\multirow{2}{*}{\rotatebox{90}{\shortstack{Chronos2\\(small)}}}
 & lr          &  & 1.1e-4 & 4.9e-4 & 4.9e-4 & 1.1e-4 & 8.5e-4 & 8.5e-4 & 8.5e-4 & 3.9e-5 & 1.6e-4 \\
 & batch\_size &  & 128    & 128    & 128    & 128    & 64     & 64     & 64     & 64     & 64     \\
\midrule
\multirow{6}{*}{\rotatebox{90}{TimesFM-2.5}}
 & lr & & 1.6e-4 & 2.0e-4 & 1.4e-4 & 1.6e-4 & 1.6e-4 & 7.7e-5 & 7.7e-5 & 7.7e-5 & 7.7e-5 \\
 & batch\_size &  & 128    & 64     & 128    & 128    & 128    & 64     & 64     & 64     & 64     \\
 & (r, alpha)  &  & (8,16) & (8,16) & (16,32)& (8,16) & (8,16) & (16,32)& (16,32)& (16,32)& (16,32) \\
 & LoRA: {"out"} & &        & \checkmark &        &        &        &        &        &        &        \\
 & LoRA: {"out","qkv\_proj"} & &        &        & \checkmark &        &        &        &        &        &        \\
 & LoRA: {"out","qkv\_proj","ff0","ff1"} & & \checkmark &        &        & \checkmark & \checkmark & \checkmark & \checkmark & \checkmark & \checkmark \\
\bottomrule
\end{tabular}
}%

\end{threeparttable}
\begin{minipage}{0.95\linewidth}
\footnotesize
\raggedright
\noindent
Ctx denotes the input CGM context window (4/12/24 hours). Pred denotes the prediction horizon (30-min/1-hr/2-hr). For TimesFM-2.5, checkmarks indicate the selected LoRA target modules (desired\_tails) for each (context, horizon) setting; (r, alpha) lists the LoRA rank and scaling factor.
\par
\end{minipage}
\end{table}

\begin{table}
\caption{Hyperparameter configurations for all models in the non-diabetes and T2D cohort, obtained via hyperparameter optimization.}
\label{tab:T2Dhyper}
\renewcommand{\arraystretch}{1.8}
\centering
\begin{threeparttable}
\resizebox{\textwidth}{!}{%
\begin{tabular}{llllllclllll}
\toprule
 & Hyperparams &  & \multicolumn{3}{c}{Ctx: 4-hr} & \multicolumn{3}{c}{Ctx: 12-hr} & \multicolumn{3}{c}{Ctx: 24-hr} \\
\cmidrule(lr){4-6} \cmidrule(lr){7-9} \cmidrule(lr){10-12}
 &  & Pred
 & 30-min & 1-hr & 2-hr
 & 30-min & 1-hr & 2-hr
 & 30-min & 1-hr & 2-hr \\
\midrule

\multirow{5}{*}{\rotatebox{90}{LSTM}}
 & lr          &  & 6.7e-4 & 6.7e-4 & 1.7e-4 & 3.4e-4 & 3.4e-4 & 1.7e-4 & 3.4e-4 & 3.4e-4 & 3.4e-4 \\
 & batch\_size &  & 256     & 256     & 128     & 64      & 64      & 128     & 64      & 64      & 64      \\
 & hidden\_dim &  & 256     & 256     & 256     & 128     & 128     & 256     & 128     & 128     & 128     \\
 & num\_layers &  & 1       & 1       & 1       & 2       & 2       & 1       & 2       & 2       & 2       \\
 & dropout     &  & 0.282   & 0.282   & 0.293   & 0.212   & 0.212   & 0.293   & 0.212   & 0.212   & 0.212   \\
\midrule

\multirow{2}{*}{\rotatebox{90}{PatchTST}}
 & lr          &  & 1.6e-3 & 1.7e-3 & 1.6e-3 & 1.6e-3 & 1.7e-3 & 6.9e-4 & 1.7e-3 & 6.9e-4 & 1.7e-3 \\
 & batch\_size &  & 256     & 128     & 256     & 256     & 128     & 256     & 128     & 256     & 128     \\
\midrule

\multirow{2}{*}{\rotatebox{90}{\shortstack{Chronos\\(bolt-tiny)}}}
 & lr          &  & 4.2e-4 & 4.2e-4 & 7.5e-5 & 1.4e-4 & 1.8e-4 & 1.4e-4 & 4.2e-4 & 4.2e-4 & 7.5e-5 \\
 & batch\_size &  & 256     & 256     & 256     & 256     & 128     & 256     & 256     & 256     & 256     \\
\midrule

\multirow{2}{*}{\rotatebox{90}{\shortstack{Chronos\\(bolt-mini)}}}
 & lr          &  & 1.3e-5 & 4.7e-4 & 7.5e-5 & 7.5e-5 & 4.2e-4 & 1.4e-4 & 1.4e-4 & 4.2e-4 & 1.4e-4 \\
 & batch\_size &  & 256     & 128     & 256     & 256     & 256     & 256     & 256     & 256     & 256     \\
\midrule

\multirow{2}{*}{\rotatebox{90}{\shortstack{Chronos2\\(small)}}}
 & lr          &  & 3.9e-5 & 1.1e-4 & 1.5e-4 & 8.5e-4 & 1.1e-4 & 8.5e-4 & 8.5e-4 & 1.1e-4 & 8.5e-4 \\
 & batch\_size &  & 64      & 128     & 64      & 64      & 128     & 64      & 64      & 128     & 64      \\
\midrule

\multirow{6}{*}{\rotatebox{90}{TimesFM-2.5}}
 & lr          &  & 7.7e-5 & 2.0e-4 & 2.0e-4 & 7.7e-5 & 7.7e-5 & 7.7e-5 & 7.7e-5 & 2.0e-4 & 2.0e-4 \\
 & batch\_size &  & 64      & 256     & 256     & 64      & 64      & 64      & 64      & 256     & 256     \\
 & (r, alpha)  &  & (16,32) & (16,32) & (16,32) & (16,32) & (16,32) & (16,32) & (16,32) & (16,32) & (16,32) \\
 & LoRA: {"out"} & &        & \checkmark & \checkmark &        &        &        &        & \checkmark & \checkmark \\
 & LoRA: {"out","qkv\_proj"} & &        &        &        &        &        &        &        &        &        \\
 & LoRA: {"out","qkv\_proj","ff0","ff1"} & & \checkmark &        &        & \checkmark & \checkmark & \checkmark & \checkmark &        &        \\
\bottomrule
\end{tabular}
}%

\end{threeparttable}
\begin{minipage}{0.95\linewidth}
\footnotesize
\raggedright
\noindent
Ctx denotes the input CGM context window (4/12/24 hours). Pred denotes the prediction horizon (30-min/1-hr/2-hr). For TimesFM-2.5, checkmarks indicate the selected LoRA target modules (desired\_tails) for each (context, horizon) setting; (r, alpha) lists the LoRA rank and scaling factor.
\par
\end{minipage}
\end{table}

\clearpage

\begin{table}[!p]
\centering
\caption{CGM forecasting performance (mean $\pm$ SD) on the T1D cohort evaluated on the in-distribution (ID) test set.}
\label{tab:cgm_perf_T1D}

\begingroup
\scriptsize
\renewcommand{\arraystretch}{0.85}
\setlength{\tabcolsep}{2pt}
\resizebox{\textwidth}{!}{%
\begin{tabular}{
  l
  >{\centering\arraybackslash}m{2.12cm}
  ccc ccc ccc
}
\toprule
\multirow{2}{*}{} & \multirow{2}{*}{Model} &
\multicolumn{3}{c}{Ctx: 4-hr} &
\multicolumn{3}{c}{Ctx: 12-hr} &
\multicolumn{3}{c}{Ctx: 24-hr} \\
\cmidrule(lr){3-5} \cmidrule(lr){6-8} \cmidrule(lr){9-11}
& & Pred: 30-min & Pred: 1-hr & Pred: 2-hr
  & Pred: 30-min & Pred: 1-hr & Pred: 2-hr
  & Pred: 30-min & Pred: 1-hr & Pred: 2-hr \\
\midrule

% ========================= RMSE block =========================
\multirow{13}{*}[-40pt]{\rotatebox{90}{RMSE}} &
LOCF\rowstrut & 12.70 & 20.96 & 33.69 & 12.70 & 20.96 & 33.69 & 12.70 & 20.96 & 33.69\\
& AutoARIMA\rowstrut & 11.86 & 22.85 & 46.00 & 10.95 & 20.26 & 37.70 & 10.59 & 18.88 & 33.87\\
& Elastic Net\rowstrut & 10.36 & 18.32 & 30.60 & 10.28 & 18.21 & 30.66 & 10.23 & 17.98 & 30.09\\
& LSTM\rowstrut & 10.28 $\pm$ 0.09 & 18.11 $\pm$ 0.11 & 30.28 $\pm$ 0.09 & 10.25 $\pm$ 0.09 & 18.07 $\pm$ 0.10 & 30.42 $\pm$ 0.17 & 10.37 $\pm$ 0.10 & 18.30 $\pm$ 0.24 & 30.05 $\pm$ 0.23\\
& PatchTST\rowstrut & 10.38 $\pm$ 0.21 &\textbf{ 17.76 $\pm$ 0.13} & \textbf{30.00 $\pm$ 0.08} & 10.54 $\pm$ 0.24 & 17.87 $\pm$ 0.27 & 30.02 $\pm$ 0.21 & 10.52 $\pm$ 0.22 & 18.07 $\pm$ 0.21 & 29.57 $\pm$ 0.19\\
& \shortstack{Chronos-bolt-tiny \\ (Zero-shot)} & 12.21 & 20.55 & 34.24 & 11.77 & 19.89 & 33.01 & 11.29 & 19.08 & 31.26\\
& \shortstack{Chronos-bolt-tiny \\ (Finetuned)} & 10.11 $\pm$ 0.06 & 17.89 $\pm$ 0.06 & 31.06 $\pm$ 0.14 & 10.14 $\pm$ 0.08 & \textbf{17.54 $\pm$ 0.05} & \textbf{29.67 $\pm$ 0.09} & \textbf{9.95 $\pm$ 0.12} & 17.36 $\pm$ 0.08 & \textbf{29.12 $\pm$ 0.07} \\
& \shortstack{Chronos-bolt-mini \\ (Zero-shot)} & 12.12 & 20.71 & 34.62 & 11.62 & 19.85 & 33.00 & 11.26 & 18.92 & 31.34\\
& \shortstack{Chronos-bolt-mini \\ (Finetuned)} & \textbf{10.07 $\pm$ 0.05} & 17.83 $\pm$ 0.05 & 31.05 $\pm$ 0.17 & \textbf{10.05 $\pm$ 0.08} & 17.56 $\pm$ 0.19 & 29.83 $\pm$ 0.22 & 9.98 $\pm$ 0.07 & \textbf{17.27 $\pm$ 0.07} & 29.30 $\pm$ 0.15\\
& \shortstack{Chronos2-small \\ (Zero-shot)} & 11.38 & 19.79 & 34.14 & 11.19 & 18.81 & 31.54 & 10.96 & 18.38 & 30.82\\
& \shortstack{Chronos2-small \\ (Finetuned)} & 10.45 $\pm$ 0.11 & 18.43 $\pm$ 0.19 & 31.81 $\pm$ 0.40 & 10.41 $\pm$ 0.16 & 17.94 $\pm$ 0.11 & 30.35 $\pm$ 0.31 & 10.50 $\pm$ 0.11 & 17.77 $\pm$ 0.08 & 29.50 $\pm$ 0.23\\
& \shortstack{TimesFM-2.5 \\ (Zero-shot)} & 10.92 & 19.75 & 33.67 & 10.33 & 18.29 & 31.06 & 10.03 & 17.57 & 29.77\\
& \shortstack{TimesFM-2.5 \\ (Finetuned)} & 10.57 $\pm$ 0.27 & 19.52 $\pm$ 0.30 & 32.68 $\pm$ 0.37 & 10.26 $\pm$ 0.09 & 17.86 $\pm$ 0.05 & 30.77 $\pm$ 0.10 & 9.96 $\pm$ 0.04 & 17.46 $\pm$ 0.05 & 29.64 $\pm$ 0.05\\

\midrule

% ========================= MAE block =========================
\multirow{13}{*}[-40pt]{\rotatebox{90}{MAE}} &
LOCF\rowstrut & 11.23 & 18.15 & 28.97 & 11.23 & 18.15 & 28.97 & 11.23 & 18.15 & 28.97\\
& AutoARIMA\rowstrut & 10.10 & 19.09 & 38.50 & 9.36 & 16.97 & 31.61 & 9.07 & 15.87 & 28.61\\
& Elastic Net\rowstrut & 8.91 & 15.49 & 25.95 & 8.84 & 15.39 & 26.01 & 8.80 & 15.18 & 25.54\\
& LSTM\rowstrut & 8.93 $\pm$ 0.08 & 15.49 $\pm$ 0.10 & 25.88 $\pm$ 0.11 & 8.90 $\pm$ 0.09 & 15.45 $\pm$ 0.09 & 26.07 $\pm$ 0.20 & 9.04 $\pm$ 0.10 & 15.75 $\pm$ 0.26 & 25.80 $\pm$ 0.26\\
& PatchTST\rowstrut & 8.98 $\pm$ 0.19 & 15.04 $\pm$ 0.11 & \textbf{25.42 $\pm$ 0.07} & 9.15 $\pm$ 0.24 & 15.17 $\pm$ 0.26 & 25.48 $\pm$ 0.20 & 9.15 $\pm$ 0.23 & 15.39 $\pm$ 0.19 & 25.11 $\pm$ 0.20\\
& \shortstack{Chronos-bolt-tiny\\(Zero-shot)} & 10.70 & 17.55 & 29.17 & 10.36 & 17.06 & 28.16 & 9.92 & 16.38 & 26.68\\
& \shortstack{Chronos-bolt-tiny\\(Finetuned)} & 8.70 $\pm$ 0.07 & 15.06 $\pm$ 0.05 & 26.33 $\pm$ 0.17 & 8.76 $\pm$ 0.10 & \textbf{14.83 $\pm$ 0.05} & \textbf{25.12 $\pm$ 0.09} & 8.59 $\pm$ 0.12 & 14.72 $\pm$ 0.08 & \textbf{24.71 $\pm$ 0.08} \\
& \shortstack{Chronos-bolt-mini\\(Zero-shot)} & 10.59 & 17.67 & 29.52 & 10.21 & 17.01 & 28.16 & 9.90 & 16.23 & 26.72\\
& \shortstack{Chronos-bolt-mini\\(Finetuned)} & \textbf{8.66 $\pm$ 0.06} & \textbf{15.01 $\pm$ 0.06} & 26.34 $\pm$ 0.18 & \textbf{8.69 $\pm$ 0.08} & 14.87 $\pm$ 0.22 & 25.30 $\pm$ 0.23 & 8.63 $\pm$ 0.08 & \textbf{14.63 $\pm$ 0.07} & 24.93 $\pm$ 0.18\\
& \shortstack{Chronos2\_small\\(Zero-shot)} & 9.90 & 16.81 & 28.94 & 9.86 & 16.10 & 26.83 & 9.66 & 15.74 & 26.21\\
& \shortstack{Chronos2\_small\\(Finetuned)} & 9.04 $\pm$ 0.12 & 15.58 $\pm$ 0.16 & 26.84 $\pm$ 0.32 & 9.07 $\pm$ 0.17 & 15.24 $\pm$ 0.13 & 25.73 $\pm$ 0.25 & 9.18 $\pm$ 0.10 & 15.11 $\pm$ 0.10 & 24.99 $\pm$ 0.20\\
& \shortstack{TimesFM-2.5\\(Zero-shot)} & 9.38 & 16.67 & 28.44 & 8.86 & 15.41 & 26.21 & 8.62 & 14.80 & 25.11\\
& \shortstack{TimesFM-2.5\\(Finetuned)} & 9.09 $\pm$ 0.26 & 16.48 $\pm$ 0.25 & 27.59 $\pm$ 0.32 & 8.83 $\pm$ 0.10 & 15.04 $\pm$ 0.05 & 25.96 $\pm$ 0.09 & \textbf{8.56 $\pm$ 0.04} & 14.71 $\pm$ 0.05 & 25.01 $\pm$ 0.06\\

\bottomrule
\end{tabular}%
}
\endgroup

\par\medskip
\begin{minipage}{0.95\linewidth}
\footnotesize
\raggedright
\noindent
Results are reported as mean $\pm$ SD across repeated runs when applicable. Classical baselines and zero-shot foundation models are deterministic and evaluated with a single run, whereas stochastic deep learning models and fine-tuned foundation models are averaged over multiple random seeds. Lower values indicate better performance.
\par
\end{minipage}
\end{table}

\clearpage
\begin{table}[!p]
\centering
\caption{CGM forecasting performance (mean $\pm$ SD) on the Non-diabetes and T2D cohort evaluated on the in-distribution (ID) test set.}
\label{tab:cgm_perf_T2D}

\begingroup
\scriptsize
\renewcommand{\arraystretch}{0.85}
\setlength{\tabcolsep}{2pt}
\resizebox{\textwidth}{!}{%
\begin{tabular}{
  l
  >{\centering\arraybackslash}m{1.8cm}
  ccc ccc ccc
}
\toprule
\multirow{2}{*}{} & \multirow{2}{*}{Model} &
\multicolumn{3}{c}{Ctx: 4-hr} &
\multicolumn{3}{c}{Ctx: 12-hr} &
\multicolumn{3}{c}{Ctx: 24-hr} \\
\cmidrule(lr){3-5} \cmidrule(lr){6-8} \cmidrule(lr){9-11}
& & Pred: 30-min & Pred: 1-hr & Pred: 2-hr
  & Pred: 30-min & Pred: 1-hr & Pred: 2-hr
  & Pred: 30-min & Pred: 1-hr & Pred: 2-hr \\
\midrule

% ========================= RMSE block =========================
\multirow{13}{*}[-40pt]{\rotatebox{90}{RMSE}} &
LOCF\rowstrut & 8.10 & 11.53 & 15.74 & 8.10 & 11.53 & 15.74 & 8.10 & 11.53 & 15.74\\
& AutoARIMA\rowstrut & 8.31 & 13.33 & 20.28 & 7.74 & 11.61 & 15.95 & 7.50 & 11.02 & 15.22\\
& Elastic Net\rowstrut & 7.48 & 10.83 & 14.59 & 7.35 & 10.51 & 14.08 & 7.33 & 10.37 & 13.57\\
& LSTM\rowstrut & 7.64 $\pm$ 0.05 & 10.82 $\pm$ 0.08 & 14.47 $\pm$ 0.07 & 7.48 $\pm$ 0.07 & 10.49 $\pm$ 0.06 & 14.09 $\pm$ 0.07 & 7.49 $\pm$ 0.06 & 10.45 $\pm$ 0.08 & 13.54 $\pm$ 0.11\\
& PatchTST\rowstrut & \textbf{7.38 $\pm$ 0.06} & 10.51 $\pm$ 0.06 & \textbf{13.79 $\pm$ 0.07} & 7.40 $\pm$ 0.09 & 10.51 $\pm$ 0.09 & 13.62 $\pm$ 0.10 & 7.47 $\pm$ 0.12 & 10.33 $\pm$ 0.05 & 13.31 $\pm$ 0.10\\
\addlinespace
& \shortstack{Chronos-bolt-tiny\\(Zero-shot)} & 8.56 & 12.26 & 16.48 & 8.26 & 11.77 & 15.99 & 7.83 & 10.95 & 14.53\\
& \shortstack{Chronos-bolt-tiny\\(Finetuned)} & 7.40 $\pm$ 0.12 & \textbf{10.34 $\pm$ 0.06} & 13.85 $\pm$ 0.05 & 7.25 $\pm$ 0.13 & \textbf{10.17 $\pm$ 0.05} & \textbf{13.42 $\pm$ 0.02} & 7.12 $\pm$ 0.06 & 10.01 $\pm$ 0.04 & 13.03 $\pm$ 0.05\\
& \shortstack{Chronos-bolt-mini\\(Zero-shot)} & 8.60 & 12.34 & 16.91 & 8.27 & 11.78 & 16.21 & 7.86 & 10.96 & 14.71\\
& \shortstack{Chronos-bolt-mini\\(Finetuned)} &\textbf{ 7.38 $\pm$ 0.15} & 10.36 $\pm$ 0.06 & 13.84 $\pm$ 0.04 &\textbf{ 7.08 $\pm$ 0.05} & 10.18 $\pm$ 0.06 & 13.44 $\pm$ 0.03 & \textbf{7.01 $\pm$ 0.04} & \textbf{9.99 $\pm$ 0.03} & \textbf{13.01 $\pm$ 0.05} \\
& \shortstack{Chronos2\_small\\(Zero-shot)} & 7.96 & 11.28 & 15.29 & 7.71 & 10.58 & 14.11 & 7.46 & 10.18 & 13.30\\
\addlinespace
& \shortstack{Chronos2\_small\\(Finetuned)} & 7.47 $\pm$ 0.02 & 10.51 $\pm$ 0.02 & 14.11 $\pm$ 0.04 & 7.40 $\pm$ 0.08 & 10.33 $\pm$ 0.04 & 13.85 $\pm$ 0.12 & 7.26 $\pm$ 0.09 & 10.00 $\pm$ 0.04 & 13.09 $\pm$ 0.10\\
& \shortstack{TimesFM-2.5\\(Zero-shot)} & 7.82 & 11.18 & 15.12 & 7.32 & 10.54 & 14.03 & 7.13 & 10.14 & 13.41\\
& \shortstack{TimesFM-2.5\\(Finetuned)} & 7.60 $\pm$ 0.11 & 11.11 $\pm$ 0.05 & 15.02 $\pm$ 0.22 & 7.24 $\pm$ 0.03 & 10.45 $\pm$ 0.03 & 14.20 $\pm$ 0.03 & 7.14 $\pm$ 0.03 & 10.21 $\pm$ 0.03 & 13.51 $\pm$ 0.06\\

\midrule

% ========================= MAE block =========================
\multirow{13}{*}[-40pt]{\rotatebox{90}{MAE}} &
LOCF\rowstrut & 7.11 & 9.83 & 13.20 & 7.11 & 9.83 & 13.20 & 7.11 & 9.83 & 13.20\\
& AutoARIMA\rowstrut & 7.17 & 11.23 & 17.08 & 6.70 & 9.85 & 13.36 & 6.50 & 9.32 & 12.74\\
& Elastic Net\rowstrut & 6.47 & 9.21 & 12.25 & 6.37 & 8.92 & 11.83 & 6.35 & 8.77 & 11.35\\
& LSTM\rowstrut & 6.68 $\pm$ 0.05 & 9.27 $\pm$ 0.08 & 12.23 $\pm$ 0.08 & 6.55 $\pm$ 0.07 & 8.98 $\pm$ 0.07 & 11.93 $\pm$ 0.07 & 6.56 $\pm$ 0.06 & 8.93 $\pm$ 0.09 & 11.42 $\pm$ 0.15\\
& PatchTST\rowstrut & \textbf{6.42 $\pm$ 0.07} & 8.91 $\pm$ 0.07 & \textbf{11.41 $\pm$ 0.07} & 6.44 $\pm$ 0.09 & 8.91 $\pm$ 0.09 & 11.26 $\pm$ 0.09 & 6.50 $\pm$ 0.12 & 8.74 $\pm$ 0.04 & 11.05 $\pm$ 0.09\\
\addlinespace
& \shortstack{Chronos-bolt-tiny\\(Zero-shot)} & 7.54 & 10.44 & 13.75 & 7.30 & 10.06 & 13.42 & 6.88 & 9.33 & 12.12\\
& \shortstack{Chronos-bolt-tiny\\(Finetuned)} & 6.44 $\pm$ 0.13 & \textbf{8.74 $\pm$ 0.06} & 11.46 $\pm$ 0.05 & 6.31 $\pm$ 0.14 & \textbf{8.59 $\pm$ 0.05} & \textbf{11.10 $\pm$ 0.03} & 6.18 $\pm$ 0.07 & 8.45 $\pm$ 0.04 & \textbf{10.78 $\pm$ 0.05} \\
& \shortstack{Chronos-bolt-mini\\(Zero-shot)} & 7.56 & 10.49 & 14.19 & 7.28 & 10.07 & 13.61 & 6.92 & 9.32 & 12.29\\
& \shortstack{Chronos-bolt-mini\\(Finetuned)} & \textbf{6.42 $\pm$ 0.15} & 8.75 $\pm$ 0.06 & 11.45 $\pm$ 0.04 & \textbf{6.14 $\pm$ 0.05} & 8.60 $\pm$ 0.06 & 11.12 $\pm$ 0.03 &\textbf{ 6.08 $\pm$ 0.04} & \textbf{8.43 $\pm$ 0.03} & \textbf{10.78 $\pm$ 0.05} \\
& \shortstack{Chronos2\_small\\(Zero-shot)} & 6.96 & 9.60 & 12.83 & 6.78 & 9.02 & 11.80 & 6.54 & 8.67 & 11.12\\
\addlinespace
& \shortstack{Chronos2\_small\\(Finetuned)} & 6.49 $\pm$ 0.02 & 8.90 $\pm$ 0.03 & 11.73 $\pm$ 0.04 & 6.47 $\pm$ 0.09 & 8.76 $\pm$ 0.05 & 11.52 $\pm$ 0.13 & 6.34 $\pm$ 0.10 & 8.46 $\pm$ 0.04 & 10.85 $\pm$ 0.08\\
& \shortstack{TimesFM-2.5\\(Zero-shot)} & 6.80 & 9.44 & 12.59 & 6.35 & 8.90 & 11.62 & 6.17 & 8.55 & 11.09\\
& \shortstack{TimesFM-2.5\\(Finetuned)} & 6.57 $\pm$ 0.09 & 9.40 $\pm$ 0.05 & 12.55 $\pm$ 0.21 & 6.27 $\pm$ 0.03 & 8.82 $\pm$ 0.03 & 11.83 $\pm$ 0.03 & 6.18 $\pm$ 0.02 & 8.63 $\pm$ 0.03 & 11.23 $\pm$ 0.07\\

\bottomrule
\end{tabular}%
}
\endgroup

\par\medskip
\begin{minipage}{0.95\linewidth}
\footnotesize
\raggedright
\noindent
Results are reported as mean $\pm$ SD across repeated runs when applicable. Classical baselines and zero-shot foundation models are deterministic and evaluated with a single run, whereas stochastic deep learning models and fine-tuned foundation models are averaged over multiple random seeds. Lower values indicate better performance.
\par
\end{minipage}
\end{table}

\clearpage

\begin{table}[p]
\centering
\caption{ID vs. OOD \textbf{RMSE} performance and fine-tuning gains on the T1D cohort.}
\label{tab:ood_gain_T1D}

\resizebox{\textwidth}{!}{%
\begin{tabular}{llcccccccc}
\toprule
& &
\multicolumn{2}{c}{\textbf{Chronos-bolt-tiny}} &
\multicolumn{2}{c}{\textbf{Chronos-bolt-mini}} &
\multicolumn{2}{c}{\textbf{Chronos2\_small}} &
\multicolumn{2}{c}{\textbf{TimesFM-2.5}} \\
\cmidrule(lr){3-4} \cmidrule(lr){5-6} \cmidrule(lr){7-8} \cmidrule(lr){9-10}
\textbf{Ctx-Pred} & \textbf{Metric}
& \textbf{ID} & \textbf{OOD}
& \textbf{ID} & \textbf{OOD}
& \textbf{ID} & \textbf{OOD}
& \textbf{ID} & \textbf{OOD} \\
\midrule

% ===================== 4-hr =====================
\multirow{3}{*}{Ctx: 4-hr, Pred: 30-min}
& Zero-shot    & 12.21 & 12.33 & 12.12 & 12.09 & 11.38 & 11.28 & 10.92 & 10.83 \\
& Finetuned    & 10.11 & 10.06 & 10.07 & 10.02 & 10.45 & 10.37 & 10.57 & 10.49 \\
& Improve (\%) 
& \imp{17.2\%} & \imp{18.4\%}
& \imp{16.9\%} & \imp{17.1\%}
& \imp{8.1\%}  & \imp{8.1\%}
& \imp{3.2\%}  & \imp{3.2\%} \\
\midrule

\multirow{3}{*}{Ctx: 4-hr, Pred: 1-hr}
& Zero-shot    & 20.55 & 21.32 & 20.71 & 21.28 & 19.79 & 20.48 & 19.75 & 20.34 \\
& Finetuned    & 17.89 & 18.57 & 17.83 & 18.53 & 18.43 & 19.02 & 19.52 & 19.92 \\
& Improve (\%) 
& \imp{12.9\%} & \imp{12.9\%}
& \imp{13.9\%} & \imp{12.9\%}
& \imp{6.9\%}  & \imp{7.2\%}
& \imp{1.2\%}  & \imp{2.0\%} \\
\midrule

\multirow{3}{*}{Ctx: 4-hr, Pred: 2-hr}
& Zero-shot    & 34.24 & 34.93 & 34.62 & 35.46 & 34.14 & 34.91 & 33.67 & 34.30 \\
& Finetuned    & 31.06 & 31.41 & 31.05 & 31.40 & 31.81 & 32.27 & 32.68 & 32.85 \\
& Improve (\%) 
& \imp{9.3\%}  & \imp{10.1\%}
& \imp{10.3\%} & \imp{11.4\%}
& \imp{6.8\%}  & \imp{7.6\%}
& \imp{2.9\%}  & \imp{4.2\%} \\
\midrule

% ===================== 12-hr =====================
\multirow{3}{*}{Ctx: 12-hr, Pred: 30-min}
& Zero-shot    & 11.77 & 11.95 & 11.62 & 11.84 & 11.19 & 11.14 & 10.33 & 10.23 \\
& Finetuned    & 10.14 & 10.17 & 10.05 & 10.09 & 10.41 & 10.46 & 10.26 & 10.14 \\
& Improve (\%) 
& \imp{13.8\%} & \imp{14.9\%}
& \imp{13.5\%} & \imp{14.8\%}
& \imp{7.0\%}  & \imp{6.1\%}
& \imp{0.7\%}  & \imp{0.8\%} \\
\midrule

\multirow{3}{*}{Ctx: 12-hr, Pred: 1-hr}
& Zero-shot    & 19.89 & 20.72 & 19.85 & 20.60 & 18.81 & 19.48 & 18.29 & 19.18 \\
& Finetuned    & 17.54 & 18.25 & 17.56 & 18.30 & 17.94 & 18.53 & 17.86 & 18.76 \\
& Improve (\%) 
& \imp{11.8\%} & \imp{11.9\%}
& \imp{11.5\%} & \imp{11.1\%}
& \imp{4.7\%}  & \imp{4.8\%}
& \imp{2.4\%}  & \imp{2.2\%} \\
\midrule

\multirow{3}{*}{Ctx: 12-hr, Pred: 2-hr}
& Zero-shot    & 33.01 & 33.51 & 33.00 & 33.48 & 31.54 & 31.80 & 31.06 & 31.47 \\
& Finetuned    & 29.67 & 29.92 & 29.83 & 30.07 & 30.35 & 30.58 & 30.77 & 31.13 \\
& Improve (\%) 
& \imp{10.1\%} & \imp{10.7\%}
& \imp{9.6\%}  & \imp{10.2\%}
& \imp{3.8\%}  & \imp{3.8\%}
& \imp{0.9\%}  & \imp{1.1\%} \\
\midrule

% ===================== 24-hr =====================
\multirow{3}{*}{Ctx: 24-hr, Pred: 30-min}
& Zero-shot    & 11.29 & 11.40 & 11.26 & 11.38 & 10.96 & 10.96 & 10.03 & 9.88 \\
& Finetuned    & 9.95 & 9.96 & 9.98 & 10.02 & 10.50 & 10.44 & 9.96 & 9.81 \\
& Improve (\%) 
& \imp{11.9\%} & \imp{12.7\%}
& \imp{11.4\%} & \imp{12.0\%}
& \imp{4.2\%}  & \imp{4.7\%}
& \imp{0.6\%}  & \imp{0.6\%} \\
\midrule

\multirow{3}{*}{Ctx: 24-hr, Pred: 1-hr}
& Zero-shot    & 19.08 & 19.62 & 18.92 & 19.55 & 18.38 & 18.92 & 17.57 & 18.17 \\
& Finetuned    & 17.36 & 17.96 & 17.27 & 17.87 & 17.77 & 18.21 & 17.46 & 18.14 \\
& Improve (\%) 
& \imp{9.0\%} & \imp{8.5\%}
& \imp{8.7\%} & \imp{8.6\%}
& \imp{3.3\%} & \imp{3.8\%}
& \imp{0.6\%} & \imp{0.1\%} \\
\midrule

\multirow{3}{*}{Ctx: 24-hr, Pred: 2-hr}
& Zero-shot    & 31.26 & 31.56 & 31.34 & 31.50 & 30.82 & 30.99 & 29.77 & 29.74 \\
& Finetuned    & 29.12 & 29.29 & 29.30 & 29.45 & 29.50 & 29.54 & 29.64 & 29.55 \\
& Improve (\%) 
& \imp{6.9\%} & \imp{7.2\%}
& \imp{6.5\%} & \imp{6.5\%}
& \imp{4.3\%} & \imp{4.7\%}
& \imp{0.4\%} & \imp{0.6\%} \\

\bottomrule
\end{tabular}
}%
\end{table}

\clearpage

\begin{table}[p]
\centering
\caption{ID vs. OOD \textbf{RMSE} performance and fine-tuning gains on the Non-diabetes and T2D cohort.}
\label{tab:ood_gain_T2D}

\resizebox{\textwidth}{!}{%
\begin{tabular}{llcccccccc}
\toprule
& &
\multicolumn{2}{c}{\textbf{Chronos-bolt-tiny}} &
\multicolumn{2}{c}{\textbf{Chronos-bolt-mini}} &
\multicolumn{2}{c}{\textbf{Chronos2\_small}} &
\multicolumn{2}{c}{\textbf{TimesFM-2.5}} \\
\cmidrule(lr){3-4} \cmidrule(lr){5-6} \cmidrule(lr){7-8} \cmidrule(lr){9-10}
\textbf{Ctx-Pred} & \textbf{Metric}
& \textbf{ID} & \textbf{OOD}
& \textbf{ID} & \textbf{OOD}
& \textbf{ID} & \textbf{OOD}
& \textbf{ID} & \textbf{OOD} \\
\midrule

% ===================== 4-hr =====================
\multirow{3}{*}{Ctx: 4-hr, Pred: 30-min}
& Zero-shot & 8.56 & 7.97 & 8.60 & 7.94 & 7.96 & 7.40 & 7.82 & 7.24 \\
& Finetuned & 7.40 & 6.96 & 7.38 & 6.88 & 7.47 & 6.96 & 7.60 & 7.15 \\
& Improve (\%)
& \impPos{13.6\%} & \impPos{12.7\%}
& \impPos{14.2\%} & \impPos{13.3\%}
& \impPos{6.2\%}  & \impPos{6.0\%}
& \impPos{2.9\%}  & \impPos{1.2\%} \\
\midrule

\multirow{3}{*}{Ctx: 4-hr, Pred: 1-hr}
& Zero-shot & 12.26 & 12.07 & 12.34 & 12.23 & 11.28 & 11.10 & 11.18 & 11.06 \\
& Finetuned & 10.34 & 10.17 & 10.36 & 10.16 & 10.51 & 10.23 & 11.11 & 10.70 \\
& Improve (\%)
& \impPos{15.6\%} & \impPos{15.8\%}
& \impPos{16.1\%} & \impPos{16.9\%}
& \impPos{6.8\%}  & \impPos{7.8\%}
& \impPos{0.6\%}  & \impPos{3.2\%} \\
\midrule

\multirow{3}{*}{Ctx: 4-hr, Pred: 2-hr}
& Zero-shot & 16.48 & 16.55 & 16.91 & 16.86 & 15.29 & 15.19 & 15.12 & 15.14 \\
& Finetuned & 13.85 & 13.89 & 13.84 & 13.87 & 14.11 & 14.14 & 15.02 & 14.97 \\
& Improve (\%)
& \impPos{16.0\%} & \impPos{16.0\%}
& \impPos{18.2\%} & \impPos{17.8\%}
& \impPos{7.7\%}  & \impPos{6.9\%}
& \impPos{0.7\%}  & \impPos{1.1\%} \\
\midrule

% ===================== 12-hr =====================
\multirow{3}{*}{Ctx: 12-hr, Pred: 30-min}
& Zero-shot & 8.26 & 7.79 & 8.27 & 7.76 & 7.71 & 7.21 & 7.32 & 6.83 \\
& Finetuned & 7.25 & 6.79 & 7.08 & 6.64 & 7.40 & 6.90 & 7.24 & 6.78 \\
& Improve (\%)
& \impPos{12.3\%} & \impPos{12.8\%}
& \impPos{14.4\%} & \impPos{14.4\%}
& \impPos{4.0\%}  & \impPos{4.3\%}
& \impPos{1.1\%}  & \impPos{0.7\%} \\
\midrule

\multirow{3}{*}{Ctx: 12-hr, Pred: 1-hr}
& Zero-shot & 11.77 & 11.62 & 11.78 & 11.60 & 10.58 & 10.44 & 10.54 & 10.29 \\
& Finetuned & 10.17 & 9.95 & 10.18 & 9.94 & 10.33 & 10.09 & 10.45 & 10.30 \\
& Improve (\%)
& \impPos{13.6\%} & \impPos{14.3\%}
& \impPos{13.6\%} & \impPos{14.3\%}
& \impPos{2.3\%}  & \impPos{3.3\%}
& \impNeg{-0.9\%} & \impNeg{-0.1\%} \\
\midrule

\multirow{3}{*}{Ctx: 12-hr, Pred: 2-hr}
& Zero-shot & 15.99 & 16.27 & 16.21 & 16.43 & 14.11 & 14.26 & 14.03 & 14.09 \\
& Finetuned & 13.42 & 13.53 & 13.44 & 13.53 & 13.85 & 13.93 & 14.20 & 14.20 \\
& Improve (\%)
& \impPos{16.1\%} & \impPos{16.8\%}
& \impPos{17.1\%} & \impPos{17.7\%}
& \impPos{1.9\%}  & \impPos{2.3\%}
& \impNeg{-1.2\%} & \impNeg{-0.8\%} \\
\midrule

% ===================== 24-hr =====================
\multirow{3}{*}{Ctx: 24-hr, Pred: 30-min}
& Zero-shot & 7.83 & 7.34 & 7.86 & 7.31 & 7.46 & 7.01 & 7.13 & 6.69 \\
& Finetuned & 7.12 & 6.66 & 7.01 & 6.54 & 7.26 & 6.75 & 7.14 & 6.69 \\
& Improve (\%)
& \impPos{9.1\%}  & \impPos{9.3\%}
& \impPos{10.9\%} & \impPos{10.5\%}
& \impPos{2.6\%}  & \impPos{3.8\%}
& \impNeg{-0.1\%} & \impNeg{-0.1\%} \\
\midrule

\multirow{3}{*}{Ctx: 24-hr, Pred: 1-hr}
& Zero-shot & 10.95 & 10.71 & 10.96 & 10.64 & 10.18 & 10.11 & 10.14 & 9.98 \\
& Finetuned & 10.01 & 9.75 & 9.99 & 9.70 & 10.00 & 9.80 & 10.21 & 10.10 \\
& Improve (\%)
& \impPos{8.6\%}  & \impPos{9.0\%}
& \impPos{8.8\%}  & \impPos{8.8\%}
& \impPos{1.7\%}  & \impPos{3.1\%}
& \impNeg{-0.7\%} & \impNeg{-1.2\%} \\
\midrule

\multirow{3}{*}{Ctx: 24-hr, Pred: 2-hr}
& Zero-shot & 14.53 & 14.72 & 14.71 & 14.81 & 13.30 & 13.55 & 13.41 & 13.64 \\
& Finetuned & 13.03 & 12.99 & 13.01 & 12.95 & 13.09 & 13.19 & 13.51 & 13.71 \\
& Improve (\%)
& \impPos{10.3\%} & \impPos{11.8\%}
& \impPos{11.5\%} & \impPos{12.6\%}
& \impPos{1.6\%}  & \impPos{2.7\%}
& \impNeg{-0.7\%} & \impNeg{-0.5\%} \\

\bottomrule
\end{tabular}
}%
\end{table}

\clearpage
\begin{table}
\caption{Performance of visual backbone models trained on the Food-101 dataset.}
\label{tab:CV_models}
\centering

\begin{threeparttable}
\begin{tabular}{lccccc}
\toprule
Model & Params (M) & lr & freeze\_backbone & Top-1 Acc (\%) & Top-5 Acc (\%) \\
\midrule
ResNet18           & 11.7  & $1\mathrm{e}{-3}$ & False & 66.93 & 90.74 \\
ResNet50           & 25.6  & $1\mathrm{e}{-3}$ & False & 74.09 & 94.04 \\
ResNet101          & 44.5  & $1\mathrm{e}{-3}$ & False & 75.55 & 94.08 \\
ResNet152          & 60.2  & $1\mathrm{e}{-3}$ & False & 73.70 & 93.67 \\
MobileNet-v3-Large & 5.5   & $1\mathrm{e}{-3}$ & False & 73.61 & 93.31 \\
DenseNet169        & 14.1  & $1\mathrm{e}{-3}$ & False & 76.03 & 94.37 \\
DenseNet201        & 20.0  & $1\mathrm{e}{-3}$ & False & 74.43 & 93.79 \\
ConvNeXT-tiny      & 28.6  & $1\mathrm{e}{-4}$ & False & 85.48 & 97.87 \\
ConvNeXT-base      & 88.6  & $1\mathrm{e}{-4}$ & False & \textbf{87.89} & \textbf{98.31} \\
EfficientNetV2-S   & 21.5  & $1\mathrm{e}{-4}$ & False & 85.42 & 97.61 \\
EfficientNetV2-M   & 54.1  & $1\mathrm{e}{-4}$ & False & 84.36 & 97.53 \\
EfficientNetV2-L   & 118.5 & $1\mathrm{e}{-4}$ & False & 87.68 & 98.19 \\
ViT-B-16           & 86.6  & $1\mathrm{e}{-4}$ & False & 83.25 & 96.96 \\
ViT-L-16           & 304.3 & $1\mathrm{e}{-4}$ & False & 86.02 & 97.86 \\
\bottomrule
\end{tabular}

\begin{tablenotes}[flushleft]
\item All visual backbone models were trained and evaluated on the Food-101 dataset. Multiple backbone fine-tuning strategies were explored in preliminary experiments. Based on validation performance, full end-to-end fine-tuning was selected and applied consistently across all models. Classification accuracy on Food-101 was used to assess representation quality, and the backbone achieving the best performance was selected to extract food image embeddings for downstream multimodal CGM forecasting. Implementation details and code are available at:
\url{https://github.com/CocoChengtw/food101-cv-models.git}
\end{tablenotes}
\end{threeparttable}
\end{table}

\begin{table}[htbp]
\centering
\caption{Leave-one-subject-out evaluation of multimodal CGM forecasting using Chronos as the CGM backbone.}
\label{tab:loso_multimodal}
\renewcommand{\arraystretch}{1.15}
\begin{tabular}{lc}
\toprule
Model configuration & LOSO RMSE \\
\midrule
CGM only & 17.5 \\
+ Image (category-supervised) & 17.0 \\
+ Image (nutrition-aligned) & 17.0 \\
+ Nutrition & 17.0 \\
+ Image (category-supervised) + Nutrition & 16.9 \\
+ Image (nutrition-aligned) + Nutrition & 16.9 \\
\bottomrule
\end{tabular}
\end{table}

\begin{table}[htbp]
\centering
\caption{Postprandial RMSE stratified by meal type for different dietary modality configurations.}
\label{tab:meal_type_multimodal}
\renewcommand{\arraystretch}{1.15}
\begin{tabular}{lccc}
\toprule
Meal type & Image & Nutrition & Image + Nutrition \\
\midrule
Breakfast & 25.6 & 24.8 & 24.5 \\
Lunch     & 25.2 & 24.7 & 24.5 \\
Dinner    & 23.5 & 24.3 & 24.0 \\
\bottomrule
\end{tabular}
\end{table}

\end{document}